\documentclass[10pt,journal,letterpaper,compsoc]{IEEEtran}

\usepackage{epsfig}
\usepackage{graphicx}
\usepackage{amsmath}
\usepackage{amssymb}
\usepackage{algorithmic}
\usepackage{algorithm}
\usepackage{subfigure}
\usepackage{hyperref}
\usepackage[applemac]{inputenc}
\usepackage{pst-sigsys}

\providecommand{\argmin}{\mathop{\textup{argmin}}}

\newcommand{\cP}{{\cal P}}

\newcommand{\om}{\omega}
\newcommand{\ga}{\gamma}
\newcommand{\la}{\lambda}
\newcommand{\LD}{{\bf L}^2({{\mathbb{R}}}^2)}
\newcommand{\LU}{{\bf L}^1({{\mathbb{R}}}^2)}

\newcommand{\V}{{\bf V}}
\newcommand{\R}{{\mathbb{R}}}

\newcommand{\Z}{{\mathbb{Z}}}

\newcommand{\N}{{\mathbb{N}}}

\title{{Invariant Scattering Convolution Networks}}
\author {Joan Bruna and St\'ephane Mallat\\
{\it CMAP, Ecole Polytechnique, Palaiseau, France}\footnote{This work is funded by the French ANR grant BLAN 0126 01.}}

\begin{document}

\maketitle


\begin{abstract}
A wavelet
scattering network computes a translation invariant image
representation, which is stable to deformations and
preserves high frequency information for classification.
It cascades wavelet transform convolutions with non-linear
modulus and averaging operators. The first network layer outputs 
SIFT-type descriptors whereas the next layers provide complementary invariant
information which improves classification. The mathematical analysis
of wavelet scattering networks explain important properties of deep
convolution networks for classification.

A scattering representation of stationary
processes incorporates higher order moments and can thus discriminate
textures having same Fourier power spectrum. 
State of the art classification results are obtained 
for handwritten digits and texture discrimination, 
with a Gaussian kernel SVM and a generative PCA classifier.
\end{abstract}

\section{Introduction}
A major difficulty of image classification comes from the
considerable variability within image classes and the inability
of Euclidean distances to measure image similarities.  
Part of this variability is due to rigid
translations, rotations or scaling.
This variability is often uninformative for
classification and should thus be eliminated.
In the framework of kernel classifiers \cite{Scholkopf}, metrics are defined
as a Euclidean distance applied on a representation $\Phi(x)$ of signals $x$. 
The operator $\Phi$ must therefore
be invariant to these rigid transformations.

Non-rigid deformations also induce important
variability within object classes  \cite{bajcsy,mnist_deformation,trouve}. 
For instance, in handwritten digit recognition, one must take into
account digit deformations due to different writing styles.
However, a full deformation invariance would reduce discrimination since 
a digit can be deformed into a different digit, for example a one into a seven.
The representation must therefore not be 
deformation invariant but continuous to deformations, to handle 
small deformations with a kernel classifier. A small deformation
of an image $x$ into $x'$ 
should correspond to a small Euclidean distance $\|\Phi(x) - \Phi(x')\|$
in the representation space, as further explained 
in Section \ref{seconinvardef}.

Translation invariant representations can be constructed
with registration algorithms \cite{Soatto} or with the Fourier transform 
modulus. However, 
Section \ref{transdef} explains why these invariants are not stable
to deformations and hence not adapted to image classification.
Trying to avoid Fourier transform instabilities 
suggests replacing sinusoidal waves by localized waveforms such as
wavelets. However, wavelet transforms are not invariant to 
translations. Building invariant representations from wavelet coefficients
requires introducing 
non-linear operators, which leads to a convolution network architecture. 

Deep convolution networks have the ability to build large-scale 
invariants which are stable to deformations  \cite{LeCun}. 
They have been applied to 
a wide range of image classification tasks.
Despite the remarkable 
successes of this neural network architecture, the properties and optimal
configurations of these networks are not well understood because 
of cascaded non-linearities.
Why use multiple layers ? How many layers ? How to optimize filters and
pooling non-linearities ?
How many internal and output neurons ? These questions are mostly 
answered through numerical experimentations that require significant
expertise. 

Deformation stability is obtained with localized wavelet
filters which separate the image variations at multiple scales and 
orientations \cite{mallat}. Computing a non-zero translation invariant
representation from wavelet coefficients requires 
introducing a non-linearity,
which is chosen to be a modulus to optimize stability \cite{Joan}.
Wavelet scattering networks,
introduced in \cite{stephane0,mallat}, build 
translation invariant representations 
with average poolings of wavelet modulus coefficients.
The output of the first network layer is similar to
SIFT \cite{SIFT} or Daisy \cite{DAISY} type descriptors.
However, this limited set of locally invariant coefficients
is not sufficiently informative to discriminate complex structures over
large-size domains. 
The information lost by the averaging is recovered by computing
a next layer of invariant coefficients, with the same wavelet convolutions and
average modulus poolings. A wavelet scattering is 
thus a deep convolution network which
cascades wavelet transforms and modulus operators.
The mathematical properties of
scattering operators \cite{mallat}
explain how these deep network coefficients relate to
image sparsity and geometry.
The network architecture is optimized in Section \ref{propscasnf}, to
retain important information while avoiding useless computations.

A scattering representation of stationary processes is introduced 
for texture discrimination. As opposed to the Fourier power spectrum, it
provides information on higher order moments and can thus discriminate
non-Gaussian textures having the same power spectrum.
Classification applications are studied in 
Section \ref{secaffi}. Scattering classification properties are demonstrated
with a Gaussian kernel SVM and 
a generative 
classifier, which selects affine space models computed with a PCA.
State-of-the-art results are obtained for handwritten digit recognition
on MNIST and USPS databes, and for texture discrimination.
Software is available at {\it www.cmap.polytechnique.fr/scattering}.

\section{Towards a Convolution Network}
\label{seconinvardef}
Section \ref{transdef} formalizes the
deformation stability condition as a Lipschitz continuity property,
and explains  why high Fourier frequencies are source of unstabilites.
Section \ref{scatwave} introduces a wavelet-based scattering transform, 
which is translation invariant and stable to deformations, and
section \ref{convnetwosec} describes its convolutional network
architecture.

\subsection{Fourier and Registration Invariants}
\label{transdef}

A representation $\Phi(x)$ is invariant to global translations 
$L_c x(u) = x(u-c)$ by $c = (c_1,c_2) \in \R^2$ if 
\begin{equation}
\label{invar}
\Phi(L_c x) = \Phi (x)~.
\end{equation}
A canonical invariant \cite{mnist_deformation,Soatto}
$\Phi(x) = x(u -a(x))$ registers $x$ with 
an anchor point $a(x)$, which is translated when $x$ is translated:
$a(L_c x) = a(x) + c$. It is therefore invariant:
$\Phi(L_c x) = \Phi(x)$.
For example, the anchor point may be a filtered
maxima $a(x) = \arg \max_{u} |x \star h(u)|$,
for some filter $h(u)$.

The Fourier transform modulus is another example of translation invariant
representation. Let  $\hat x (\omega)$ be the Fourier transform of $x(u)$.
Since  
$\widehat {L_c x} (\omega) = e^{-i c.\omega}\, \hat x(\omega)$, it results that
$|\widehat {L_c x} | = |\hat x|$ does not depend upon $c$.

To obtain appropriate similarity measurements between images which
have undergone non-rigid transformations, the representation must 
also be stable to small deformations. A small deformation can be
written $L_\tau x (u) = x(u-\tau(u))$ where $\tau(u)$ depends upon $u$ and thus
deforms the image. The deformation gradient tensor
$\nabla \tau (u)$ is a matrix whose
norm $|\nabla \tau (u) |$ measures the deformation amplitude at $u$.
A small deformation is an invertible transformation if
$|\nabla \tau (u)| < 1$
\cite{allassoniere, trouve}.
Stability to deformations is expressed as a 
Lipschitz continuity condition relative to this deformation metric:
\begin{equation}
\label{elansfl}
\|\Phi(L_\tau x) - \Phi (x) \| 
\leq C\, \|x\|\, \sup_u |\nabla \tau (u)|~,
\end{equation}
where $\|x\|^2 = \int |x(u)|^2 \, du$.
This property implies global translation invariance, because if
$\tau (u) = c$ then $\nabla \tau(u) = 0$, but it is much stronger.

A  Fourier modulus is translation invariant but unstable 
with respect to
deformations at high frequencies.
Indeed, $|\,|\hat x (\omega)| - |\widehat {L_\tau x} (\om)|\,|$
can be arbitrarily large at a high frequency $\omega$, even
for small deformations and in particular small dilations. 
As a result, $\Phi(x) = |\hat x|$ does not satisfy the
deformation continuity condition (\ref{elansfl}) \cite{mallat}.
A Fourier modulus also loses too much information.
For example,
a Dirac $\delta (u)$ and a linear chirp $e^{i u^2}$ are totally
different signals having Fourier
transforms whose moduli are equal and constant. Very different signals
may not be discriminated from their Fourier modulus.

A registration invariant $\Phi(x) = x(u - a(x))$ carries more information
than a Fourier modulus, and characterizes $x$ up to 
a global absolute position information \cite{Soatto}.
However, it has the same high-frequency instability
as a Fourier transform. Indeed, for any choice of anchor
point $a(x)$, applying
the Plancherel formula proves that
\begin{equation}
\label{Fourregis}
\| x(u-a(x)) - x'(u-a(x')) \|\geq (2 \pi)^{-1}\,\||\hat x(\om)| - |\hat x'(\om)| \| ~.
\end{equation}
If $x' = L_\tau x$, 
the Fourier transform instability at high frequencies implies that
$\Phi(x) = x(u-a(x))$ is also unstable with respect to deformations.

\subsection{Scattering Wavelets}
\label{scatwave}

A wavelet is a localized waveform and is thus stable to deformation,
as opposed to the Fourier sinusoidal waves. A wavelet transform computes
convolutions with wavelets. It is thus
translation covariant, not invariant. 
A scattering transform
computes non-linear invariants with modulus and averaging pooling functions.

Two-dimensional directional wavelets are obtained by scaling and rotating
a single band-pass filter $\psi$.
Let $G$ be a discrete, finite rotation group in $\R^2$.
Multiscale directional wavelet filters are defined for any $j \in \Z$
and rotation $r \in G$ by
\begin{equation}
\label{rotdilwave}
\psi_{2^jr} (u) =  2^{2j} \psi (2^{j} r^{-1} u)~.
\end{equation}
If the Fourier transform $\hat \psi(\omega)$ is centered at a frequency
$\eta$ then $\hat \psi_{2^j r} (\omega) = \hat \psi(2^{-j} r^{-1} \omega)$
has a support centered at
$2^j r \eta$, with a bandwidth proportional to $2^j$. 
To simplify notations, we denote $\lambda = 2^j r \in \Lambda=G \times \Z$, 
and $|\lambda| = 2^j$.

A wavelet transform filters $x$ using a family of 
wavelets: $\{x \star \psi_\la (u) \}_{\la}$. It is computed with a filter
bank of dilated and rotated wavelets having no orthogonality
property. As further explained in
Section \ref{enerinvsf}, it is stable and invertible if 
the rotated and scaled wavelet filters
cover the whole frequency plane.
On discrete images, to avoid aliasing, we only
capture frequencies in the circle $|\om| \leq \pi$ inscribed in the 
image frequency square. However, most digital natural images and textures 
have negligible energy outside this frequency circle.

\setcounter{subfigure}{0}
\begin{figure*}[ht]
\centering
\subfigure[ ]{
\includegraphics[scale=0.25]{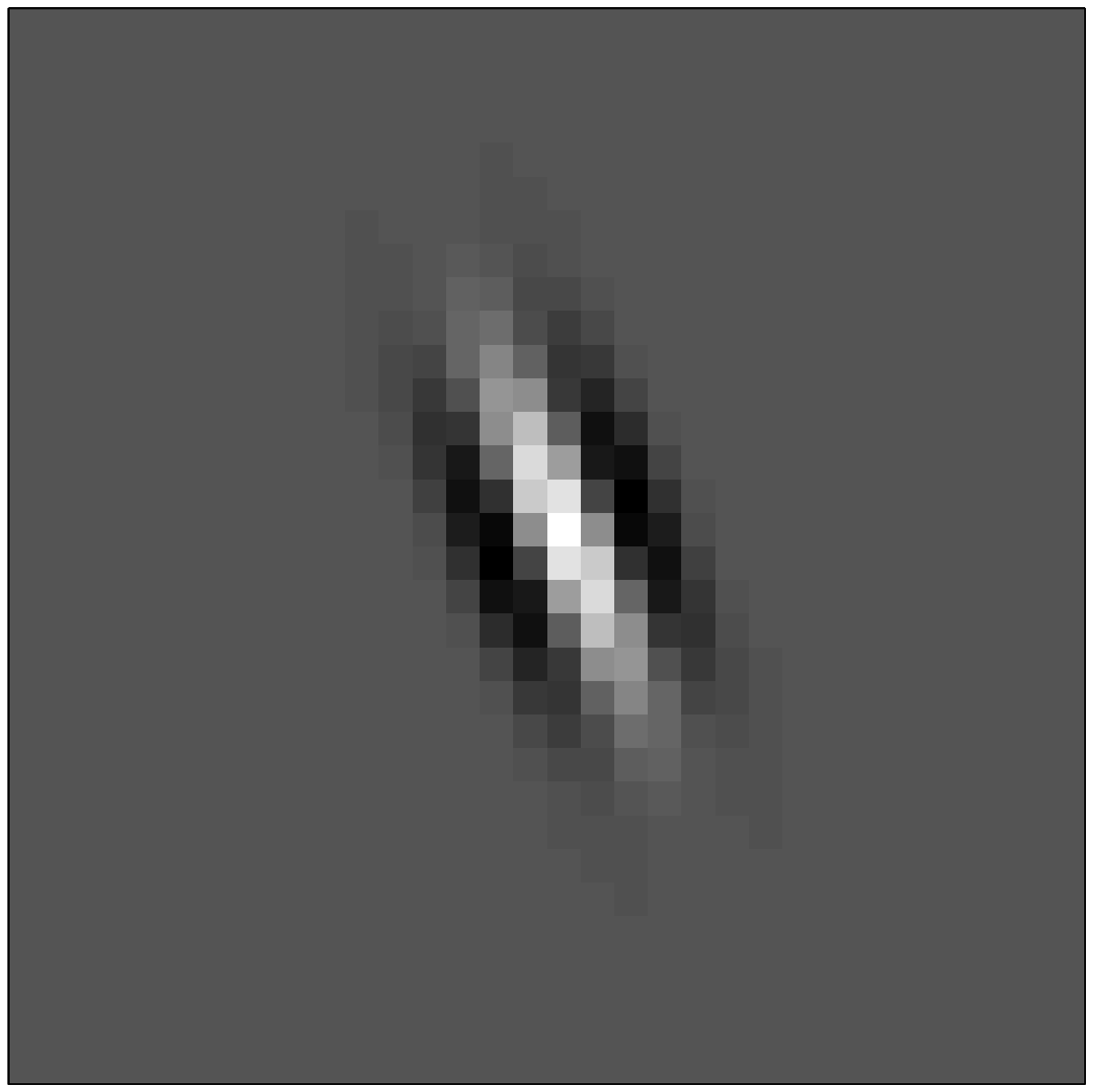}
\label{morlet_space}
}
\subfigure[ ]{
\includegraphics[scale=0.25]{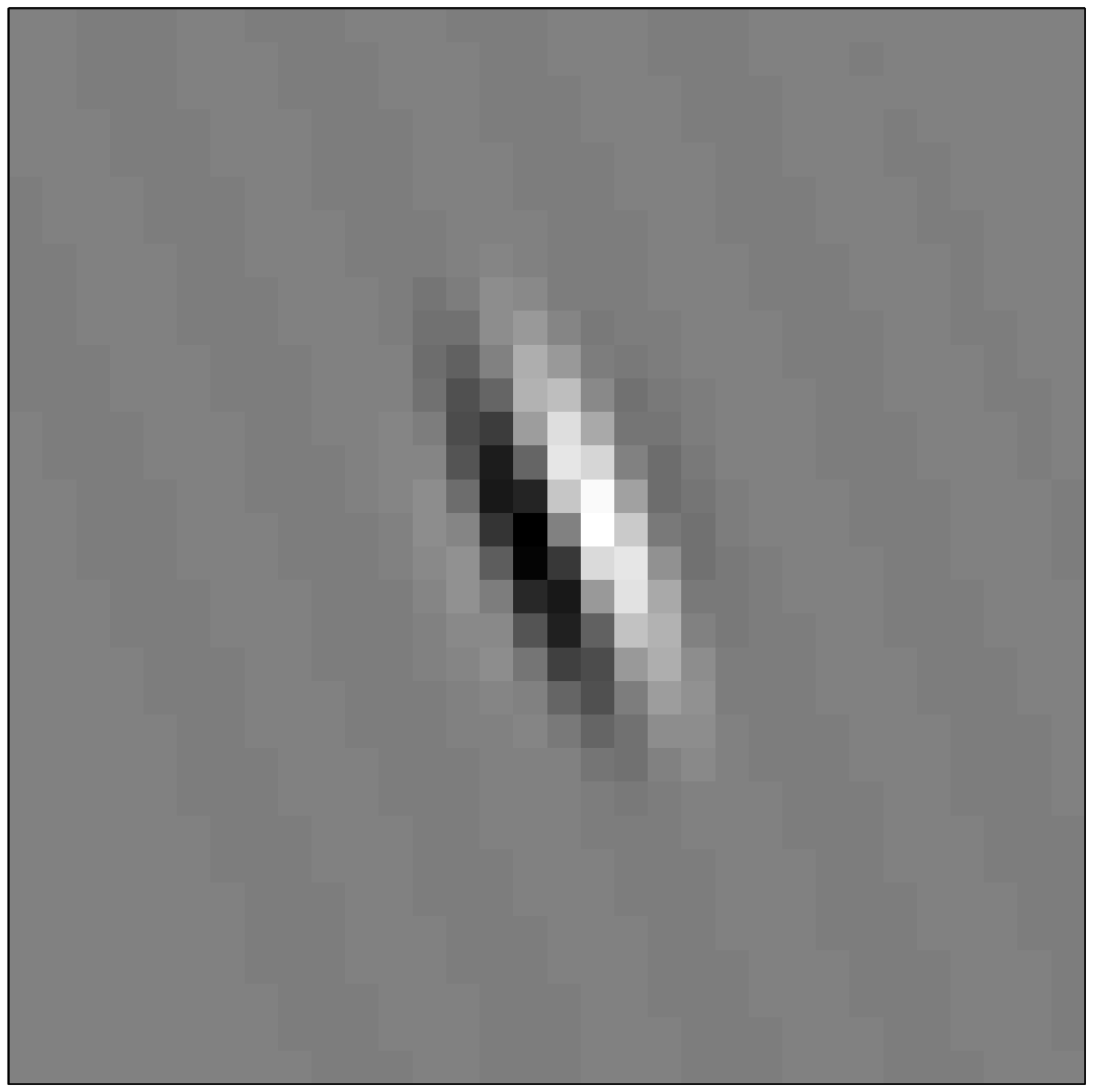}
\label{morlet_imag}
}
\subfigure[ ]{
\includegraphics[scale=0.25]{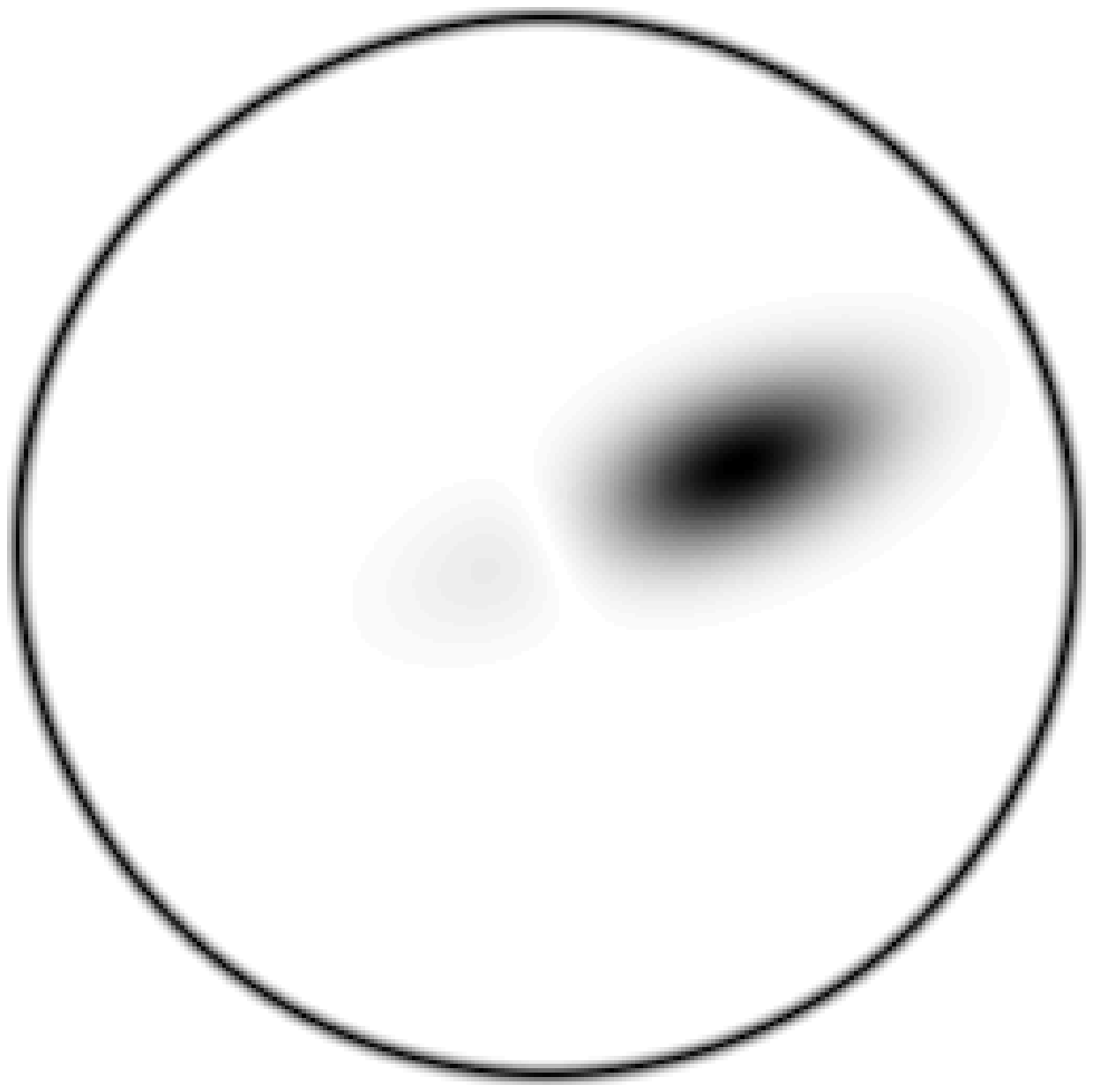}
\label{morlet_fourier}
}
\caption{Complex Morlet wavelet. (a): 
Real part of $\psi$. (b): Imaginary part of $\psi$. 
(c): Fourier modulus $|\hat{\psi}|$.
}
\label{waveletfig}
\end{figure*}

Let $u.u'$ and $|u|$ denote the inner product and norm in $\R^2$.
A Morlet wavelet $\psi$ is an example of wavelet given by
\[
\psi (u) =  C_1\, (e^{i u. \xi } -C_2)\, e^{- |u|^2/(2 \sigma^2)}~,
\]
where $C_2$ is adjusted so that
$\int \psi(u)\,du = 0$.
Figure \ref{waveletfig} shows the Morlet wavelet
with $\sigma=0.85$ and $\xi=3\pi/4$, used in
all classification experiments.

A wavelet transform commutes with 
translations, and is therefore not translation invariant. 
To build a translation invariant representation,
it is necessary to introduce a non-linearity.
If $R$ is a linear or non-linear operator which commutes with translations,
$R (L_c x) = L_c R x$, then the integral
$\int R x(u)\,du$ is translation invariant.
Applying this to $R x = x \star \psi_\la$ gives a trivial invariant
$\int x \star \psi_\lambda (u)\,du = 0$ for all $x$
because $\int \psi_\lambda (u)\,du = 0$.
If $R x = M(x \star \psi_\la)$ but $M$ is linear 
and commutes with translations then the integral still 
vanishes, which imposes choosing a non-linear $M$.
Taking advantage of the wavelet transform stability to deformations,
to obtain integrals which are also stable to deformations we also impose
that $M$ commutes with deformations
\[
\forall \tau(u)~~,~~M \,L_\tau = L_\tau\, M~.
\]
By adding a weak differentiability condition, one can prove
\cite{Joan} that $M$ must necessarily be a pointwise operator, which means
that $M x(u)$ only depends on the value $x(u)$. If we also impose
an $\LD$ stability
\[
\forall (x,y) \in \LD^2~,\, \|M x \| = \|x\|~~\mbox{and}~~
\|M x - M y \| \leq \|x - y \|,
\]
then one can verify \cite{Joan} that necessarily
$M x = e ^{i \alpha}\, |x|$, and we set $\alpha = 0$.
The resulting translation invariant coefficients are therefore
$\LU$ norms: $\|x \star \psi_\la \|_1 = \int |x \star \psi_\lambda (u)|\,du $.

The $\LU$ norms $\{ \|x \star \psi_\la \|_1 \}_\lambda$
form a crude signal representation, which measures the sparsity of 
the wavelet coefficients. For appropriate wavelets,
one can prove \cite{waldspurger} that
$x$ can be reconstructed 
from $\{ |x \star \psi_\lambda (u)| \}_\la$, up to a multiplicative constant.
The information loss thus comes from 
the integration of $|x \star \psi_{\lambda} (u)|$,
which removes all non-zero frequency components.
These non-zero frequencies can be recovered
by calculating the wavelet coefficients
$\{|x \star \psi_{\lambda_1} | \star \psi_{\lambda_2} (u)\}_{\lambda_2}$
of $|x \star \psi_{\lambda_1} |$. 
Their $\LU$ norms define a much larger family of invariants, 
for all $\lambda_1$ and $\lambda_2$:
\[
\||x \star \psi_{\la_1}| \star \psi_{\la_2} \|_1 = 
\int ||x \star \psi_{\la_1} (u)| \star \psi_{\la_2}|\,du \,.
\]

More translation invariant coefficients can be computed
by further iterating on the wavelet transform and
modulus operators. Let $U[\la] x = |x \star \psi_\la|$. 
Any sequence $p = (\la_1,\la_2,...,\la_m)$ defines a 
\emph{path}, i.e, the ordered product of non-linear 
and non-commuting operators
\[
U[p] x = U[\la_m]\,...\,U[\la_2]\,U[\la_1] x =
|~||x \star \psi_{\la_1}| \star \psi_{\la_2}|\,...\,| \star \psi_{\la_m}|~,
\]
with $U[\emptyset] x = x$.
A scattering transform along the path $p$ is 
defined as an integral, normalized by the response of a Dirac:
\[
\overline S x(p) = \mu_p^{-1}\, \int U[p] x(u)\,du~~\mbox{with}~~
\mu_p = \int U[p] \delta(u)\,du~.
\]
Each scattering coefficient $\overline S x(p)$ is invariant to a
translation of $x$. 
We shall see that this transform has many similarities with the 
Fourier transform modulus, which is also translation invariant. However,
a scattering is Lipschitz continuous to deformations 
as opposed to the Fourier transform modulus.

For classification, it is often better to compute localized descriptors 
which are invariant to translations smaller than a predefined scale $2^J$, 
while keeping the spatial variability at scales larger than $2^J$. 
This is obtained by localizing the scattering integral with a scaled
spatial window $\phi_{2^J} (u) = 2^{-2J} \phi(2^{-J} u)$.
It defines a windowed scattering transform
in the neighborhood of $u$:
\[
S_J [p] x(u) = U[p] x \star \phi_{2^J} (u) = \int U[p] x(v) \phi_{2^J} (u-v)\,dv~,
\]
and hence
\[
S_J [p] x(u) = |~||x \star \psi_{\la_1}| \star \psi_{\la_2}|\,...\,| \star \psi_{\la_m}| \star \phi_{2^J}(u)~,
\]
with $S_J [\emptyset] x = x \star \phi_{2^J}$.
For each path $p$, $S_J [p] x(u)$ is a function of 
the window position $u$, which
can be subsampled at intervals proportional to the window size $2^J$.
The averaging by $\phi_{2^J}$ implies that $S_J [p] x(u)$ 
is nearly invariant
to translations $L_c x (u) = x(u-c)$ if  $|c| \ll 2^J$. 
Section \ref{enerinvsf} proves that it is also stable 
relatively to deformations.

\subsection{Scattering Convolution Network}
\label{convnetwosec}

If $p$ is a path of length $m$ then
$S_J [p] x(u)$ is called scattering coefficient of order $m$
at the scale $2^J$. It is computed at
the layer $m$ of a convolution network which is specified.
For large scale invariants, 
several layers are necessary to avoid losing crucial
information.

For appropriate wavelets, 
first order coefficients $S_J [\lambda_1] x$ are equivalent to
SIFT coefficients \cite{SIFT}.
Indeed, SIFT computes the local sum of image gradient amplitudes among
image gradients having nearly the same direction, in
a histogram having $8$ different direction bins. 
The DAISY approximation \cite{DAISY} shows that these
coefficients are well approximated by 
$S_J [2^j r] x = |x \star \psi_{2^j r}| \star \phi_{2^J} (u)$ 
where $\psi_{2^j r}$ is the partial derivative of a Gaussian 
computed at the finest image scale $2^j$, for $8$ 
different rotations $r$. The averaging filter
$\phi_{2^J}$ is a scaled Gaussian. 

Partial derivative wavelets are well adapted to detect edges or
sharp transitions but
do not have enough frequency and directional resolution 
to discriminate complex directional structures.
For texture analysis, many researchers 
\cite{Malik,Zeevi,simoncelli} have been using 
averaged wavelet coefficient amplitudes 
$|x \star \psi_\lambda | \star \phi_J(u)$, but calculated with a
complex wavelet $\psi$ having a better frequency and directional resolution.

A scattering transform computes 
higher-order coefficients by further iterating on wavelet transforms
and modulus operators.
At a maximum scale $2^J$,
wavelet coefficients are computed at frequencies 
$2^{j} \geq 2^{-J}$, and 
lower frequencies are filtered by 
$\phi_{2^J} (u) = 2^{-2J} \phi(2^{-J} u)$. 
Since images are real-valued signals,
it is sufficient to consider ``positive'' rotations $r \in G^+$
with angles in $[0,\pi)$:
\begin{equation}
\label{wavedfn}
W_J x(u) = \Big\{
x \star \phi_{2^J} (u)~,~
x \star \psi_\lambda (u) \Big\}_{\lambda \in \Lambda_J}
\end{equation}
with $\Lambda_J = \{\lambda = 2^j r~:~r \in G^+, j \geq -J \}$.
For a Morlet wavelet $\psi$, the averaging filter $\phi$ is 
chosen to be a Gaussian.
Let us emphasize that $2^J$ is a spatial scale variable whereas
$\lambda = 2^j r$ is assimilated to a frequency variable. 

\begin{figure*}[ht]
\centering
\includegraphics[scale=1]{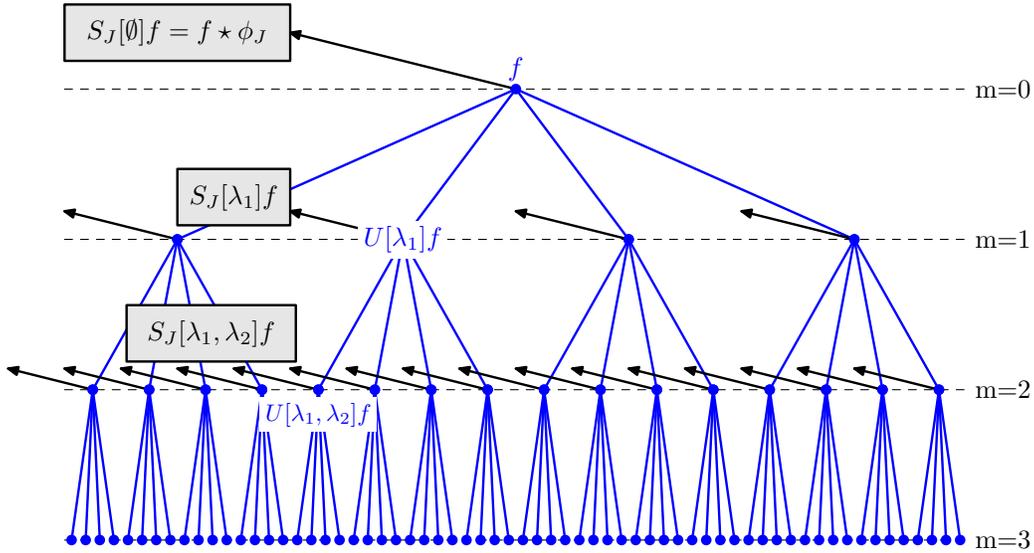}
	\caption{A scattering propagator $U_J$ applied to $x$ 
computes each
$U[\la_1] x = |x \star \psi_{\la_1}|$ and outputs 
$S_J [\emptyset] x = x \star \phi_{2^J}$ (black arrow).
Applying $U_J$ to each $U[\la_1] x$ computes all
$U[\la_1,\la_2]x$ and outputs 
$S_J[\la_1] = U[\la_1] \star \phi_{2^J}$ (black arrows).
Applying  
$U_J$ iteratively to each $U[p]x$ outputs $S_J[p]x = U[p]x \star \phi_{2^J}$
(black arrows) and computes the next path layer.
}
	\label{scattering-cascade}
\end{figure*}

A wavelet modulus propagator keeps the low-frequency
averaging and computes the modulus
of complex wavelet coefficients: 
\begin{equation}
\label{wavedfnmod}
{U}_J x(u) = \Big\{
x \star \phi_{2^J} (u)~,~
|x \star \psi_{\lambda} (u)|
\Big\}_{\lambda \in \Lambda_J}~.
\end{equation}
Let $\Lambda_J^m$ be the set of all paths $p=(\lambda_1,...,\la_m)$ 
of length $m$.
We denote
$U[\Lambda_J^m]x = \{ U [p] x \}_{p \in \Lambda_J^m}$ and 
$S_J[\Lambda_J^m]x = \{ S_J [p] x \}_{p \in \Lambda_J^m}$.
Since
\[
U_J\,U[p] x = 
\Big\{ U[p] x \star \phi_{2^J}\,,\,|U[p] x \star \psi_\lambda| \Big\}~,
\]
and $S_J[p] x = U[p] x \star \phi_{2^J}$, it results that
\begin{equation}
\label{wavedfnmod3sdf}
U_J \,U[\Lambda_J^m] x = \{U_J \,U[p] x \}_{p \in \Lambda_J^m} =
\Big\{S_J [\Lambda_J^m] x \, , \, U [\Lambda_J^{m+1}] x \Big\}~.
\end{equation}
This implies that $S_J [p] x$ can be computed along paths of length 
$m \leq m_{\max}$
by first calculating $U_J x = \{ S_J [\emptyset] x\,,\,U[\Lambda_J^1] x \}$
and iteratively applying $U_J$ to each
$U[\Lambda_J^m] x$ for increasing $m \leq m_{\max}$. This algorithm is illustrated in
Figure \ref{scattering-cascade}. 

A scattering transform thus appears to be 
a deep convolution network \cite{LeCun}, with some particularities.
As opposed to most convolution networks, a
scattering network 
outputs coefficients $S_J [p] x$ at all layers $m \leq m_{\max}$,
and not just at
the last layer $m_{\max}$ \cite{LeCun}. 
The next section proves that the energy
of the deepest layer converges quickly to zero as $m_{\max}$ increases.

A second distinction is that 
filters are not learned from data but are predefined wavelets.
Wavelets are stable with respect 
to deformations and provide sparse image representations. Stability to
deformations is a strong condition which imposes a separation of the
different image scales \cite{mallat}, hence
the use of wavelets. 

The modulus operator which recombines real and imaginary parts
can be interpreted as a {\it pooling} function in
the context of convolution networks. 
The averaging by $\phi_{2^J}$ at the output is also a pooling operator
which aggregates coefficients to build an invariant.
It has been argued \cite{midlevel}
that an average pooling loses information,
which has motivated the use of other operators such as 
hierarchical maxima \cite{Poggio}. The high frequencies lost by the
averaging are 
recovered as wavelet coefficients in the next layers, which explains the
importance of using a multilayer network structure.
As a result, it only loses the phase of these
wavelet coefficients. This phase
may however be recovered from the modulus thanks to the
wavelet transform redundancy. It has been proved 
\cite{waldspurger} that the wavelet-modulus operator
$U_J x = \{x \star \phi_{2^J}, \, |x \star \psi_\lambda| \}_{\la \in \Lambda_J}$ 
is invertible with a continuous inverse. It means that
$x$ and hence the complex phase of each $x \star \psi_\lambda$ 
can be reconstructed.
Although $U_J$ is invertible, the scattering transform is
not exactly invertible because of instabilities.
Indeed, applying $U_J$ in (\ref{wavedfnmod3sdf}) 
for $m \leq m_{\max}$ 
computes all $S_J [\Lambda_J^m] x$ for $m \leq m_{\max}$
but also the last layer of internal network coefficients
$U [\Lambda_J^{m_{\max}+1}] x$. The next section 
proves that 
$U [\Lambda_J^{m_{\max}+1}] x$ can be neglected because its
energy converges to zero as $m_{\max}$ increases. 
However, this introduces a small
error which accumulates when iterating on $U_J^{-1}$.

\begin{figure*}[ht]
\centering
\includegraphics[scale=0.25]{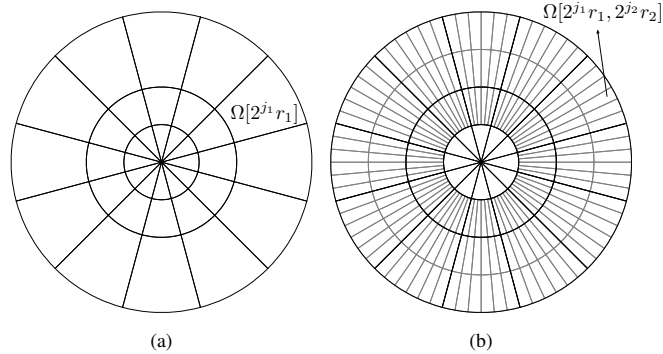}
\caption{For $m=1$ and
$m=2$, a scattering is displayed as piecewise
constant functions equal to $S_J [p] x(u)$ 
over each frequency subset $\Omega[p]$.
(a): For $m=1$, each $\Omega[2^{j_1} r_1]$ is a rotated quadrant 
of surface proportional to $2^{j_1}$.
(b): For $m=2$, each $\Omega[2^{j_1} r_1]$ is subdivided
into a partition of subsets $\Omega[2^{j_1} r_1,2^{j_2} r_2]$.}
\label{scat_display_construction}
\end{figure*}

Scattering coefficients can be displayed in the frequency plane.
Let $\{\Omega [p]\}_{p \in \Lambda_J^m}$ be a partition of $\R^2$. 
To each frequency $\om \in \R^2$ we associate the path $p(\om)$ 
such that $\omega \in \Omega [p]$. 
We display $S_J [p(\omega)] x(u)$, which is 
a piecewise constant function of $\omega \in \R^2$,
for each position $u$ and each $m=1,2$.
For $m = 1$, 
each $\Omega[{2^{j_1} r_1}]$ is chosen to be a quadrant 
rotated by $r_1$, to approximate
the frequency support of $\hat \psi_{2^{j_1} r_1}$, whose size is proportional
to $\|\psi_{2^{j_1} r_1}\|^2$ and hence
to $2^{j_1}$. This defines a partition of a dyadic annulus
illustrated in Figure \ref{scat_display_construction}(a). 
For $m=2$, $\Omega[2^{j_1} r_1 , 2^{j_2} r_2]$ is obtained by subdividing
$\Omega[2^{j_1} r_1]$,
as illustrated in Figure \ref{scat_display_construction}(b). 
Each $\Omega[2^{j_1} r_1]$ is subdivided along the 
radial axis into quadrants indexed by $j_2$.
Each of these quadrants are themselves subdivided along the angular variable
into rotated quadrants $\Omega[2^{j_1} r_1,2^{j_2} r_2]$ having a
surface proportional to  $\| |\psi_{2^{j_1} r_1}| \star \psi_{2^{j_2} r_2}\|^2$.

\setcounter{subfigure}{0}

\setcounter{subfigure}{0}
\begin{figure*}[ht]
\centering
\subfigure{
\includegraphics[scale=0.25,trim=0.0in 0 0.0in 0, clip]{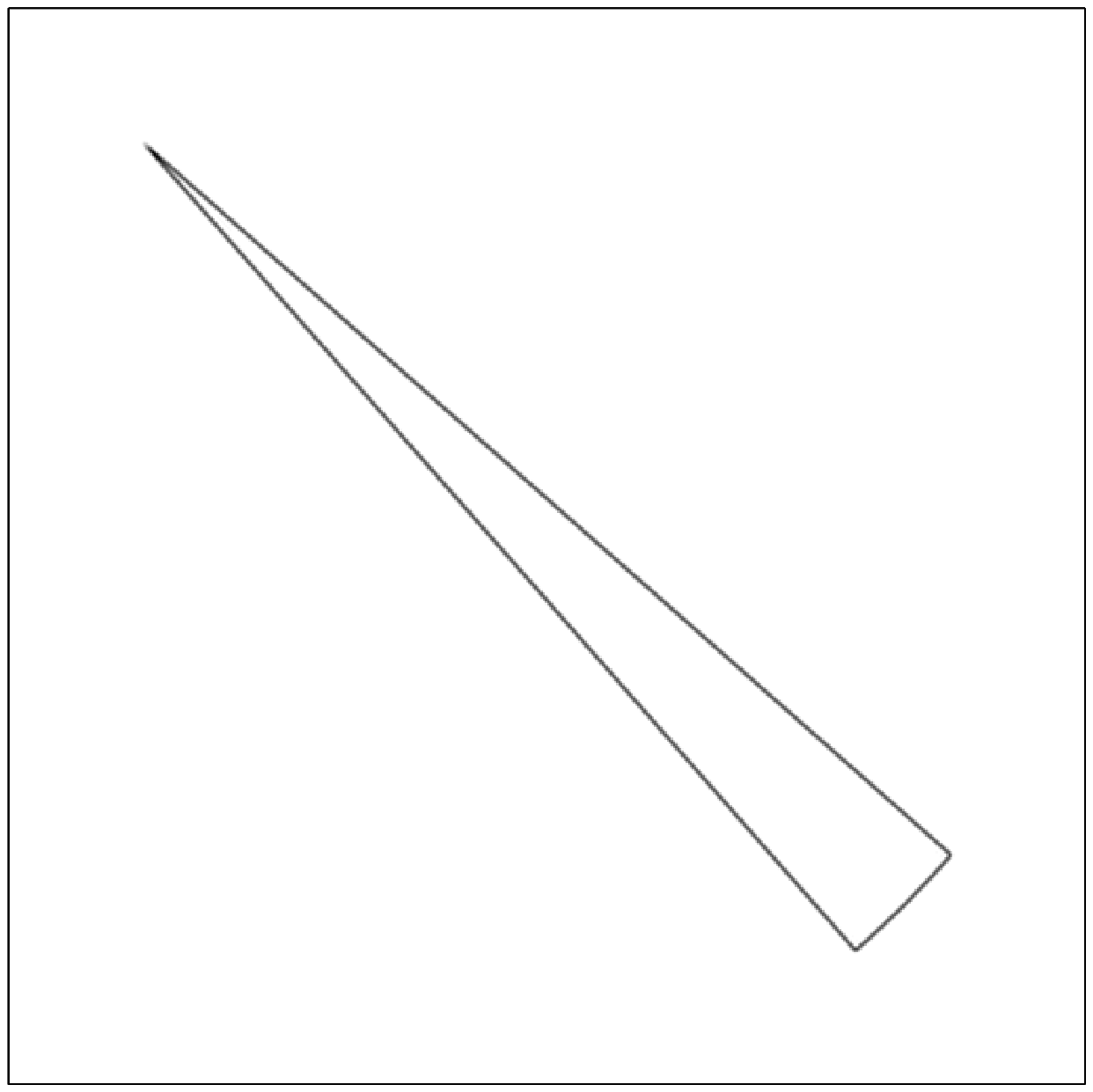}
}
\subfigure{
\includegraphics[scale=0.25,trim=0.0in 0 0.0in 0, clip]{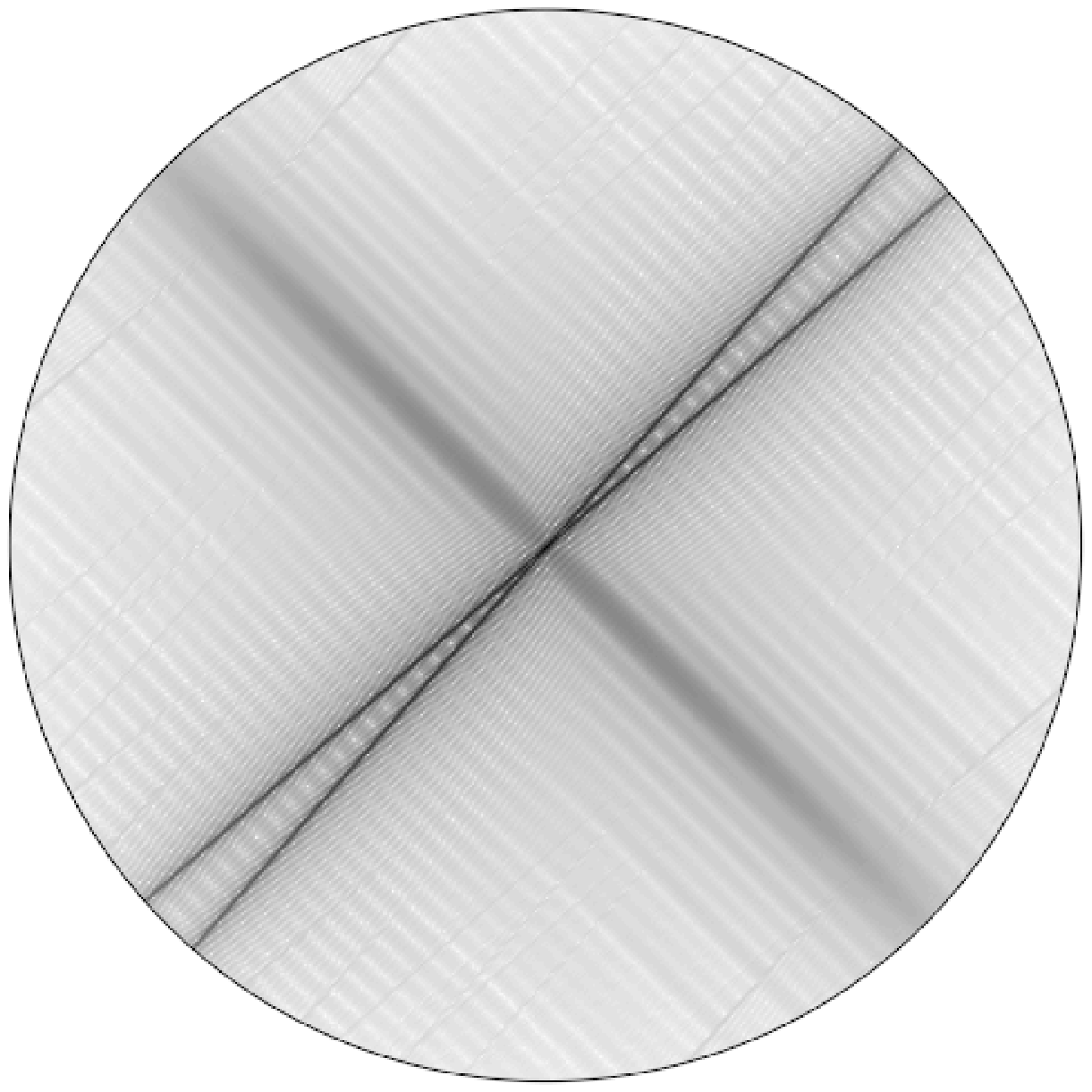}
}
\subfigure{
\includegraphics[scale=0.25,trim=0.0in 0 0.0in 0, clip]{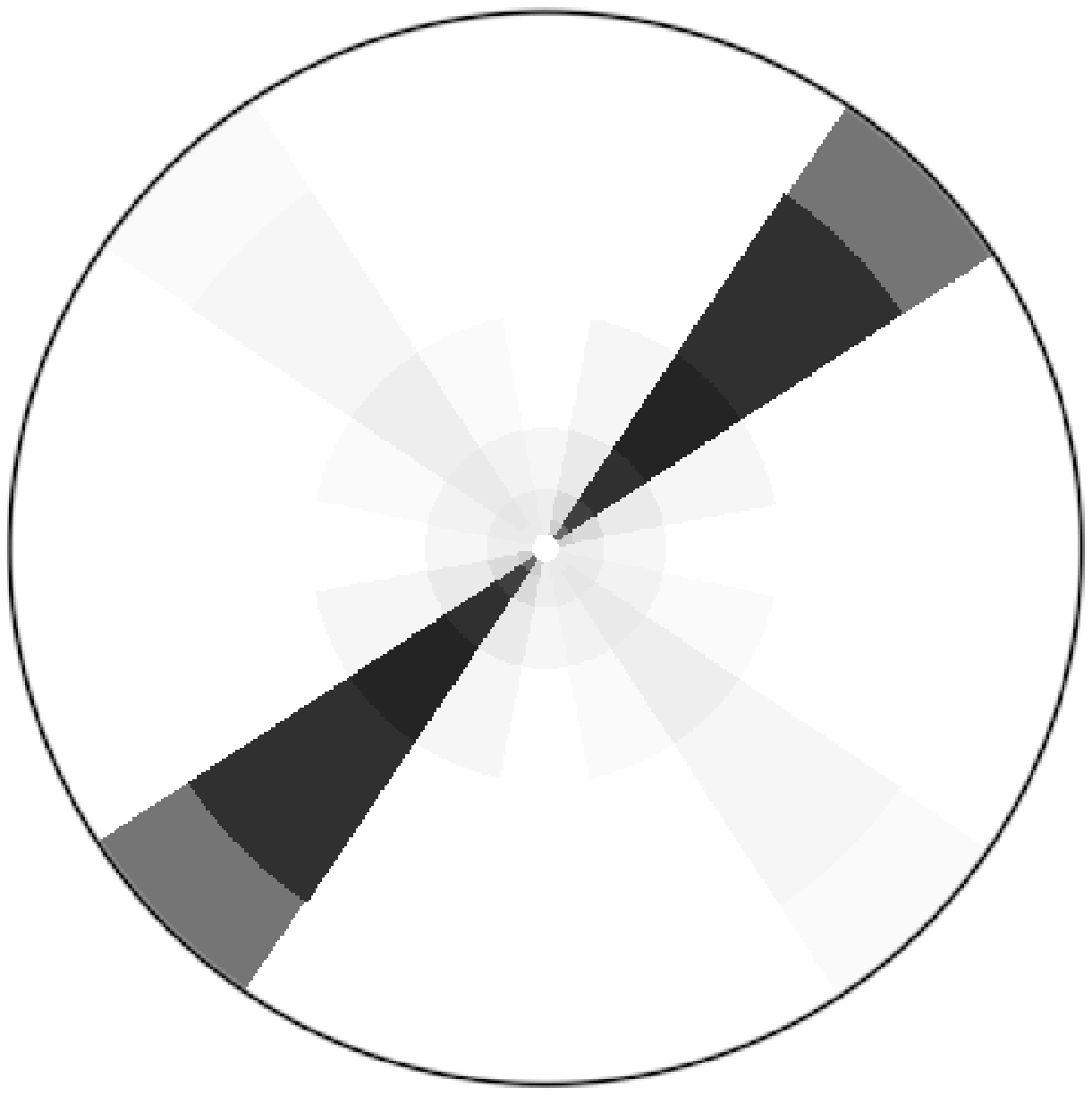}
}
\subfigure{
\includegraphics[scale=0.15,trim=0.0in 0 0.0in 0, clip]{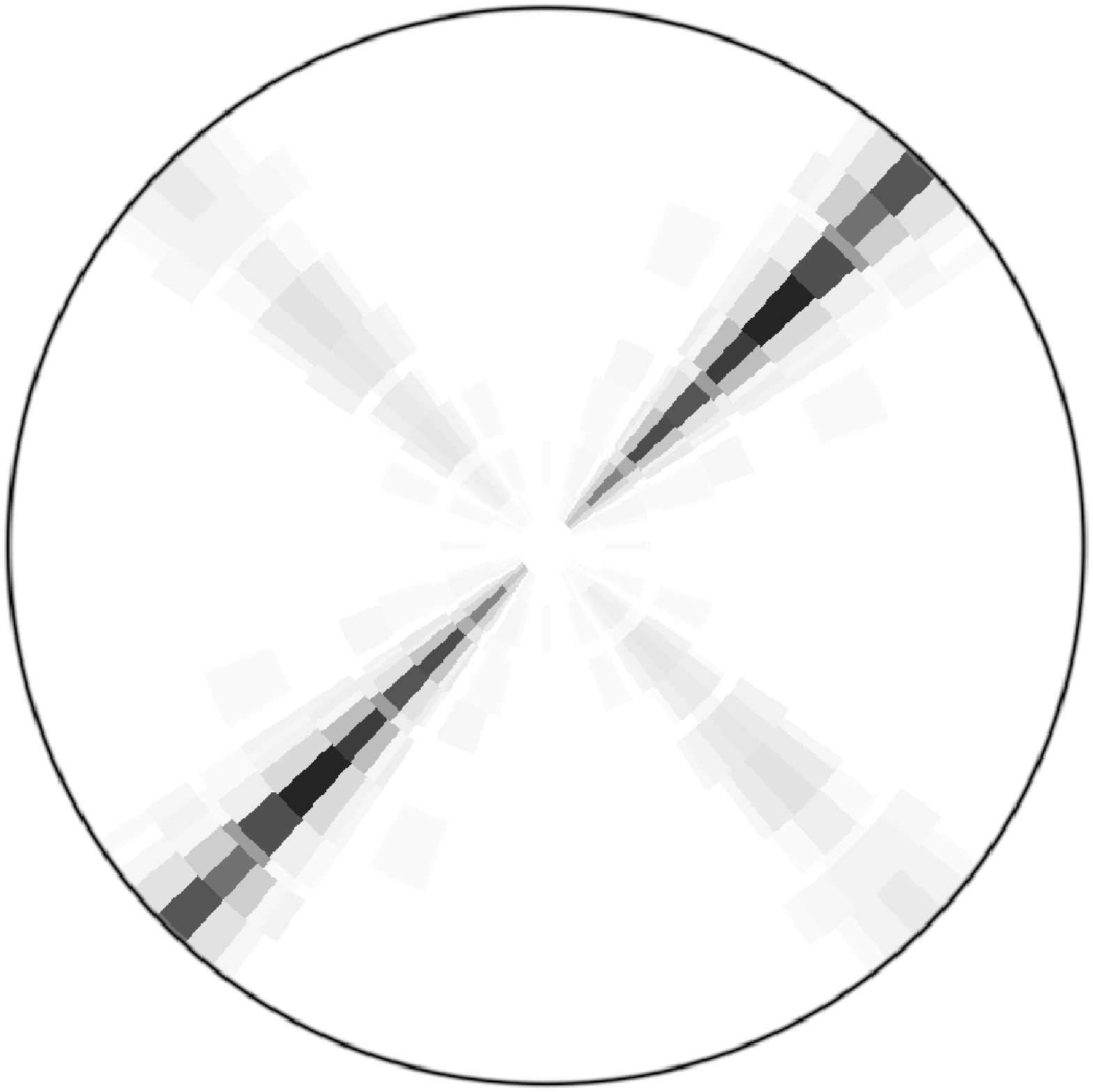}
} \\
\setcounter{subfigure}{0}
\subfigure[]{
\includegraphics[scale=0.25,trim=0.0in 0 0.0in 0, clip]{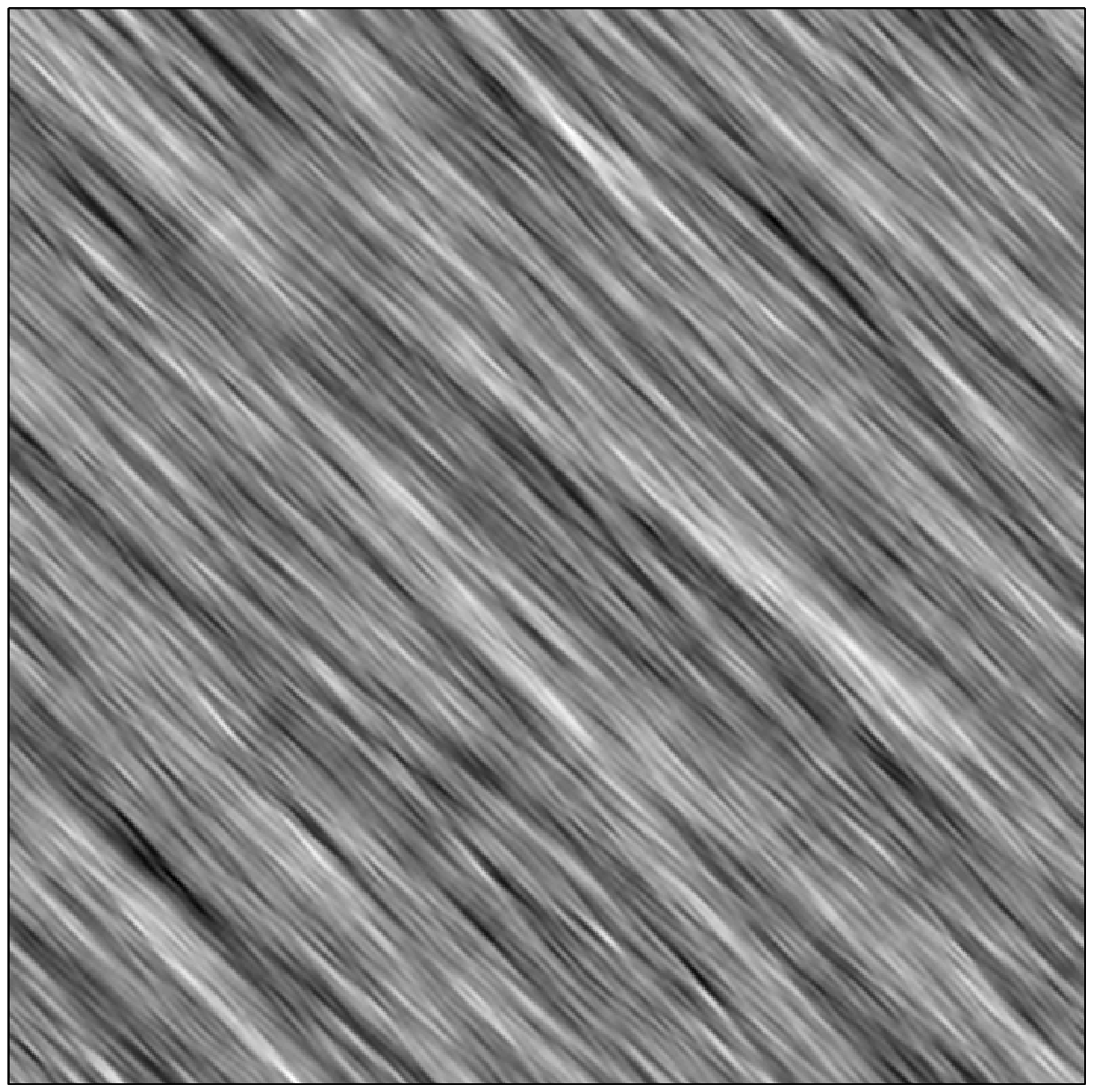}
}
\subfigure[]{
\includegraphics[scale=0.25,trim=0.0in 0 0.0in 0, clip]{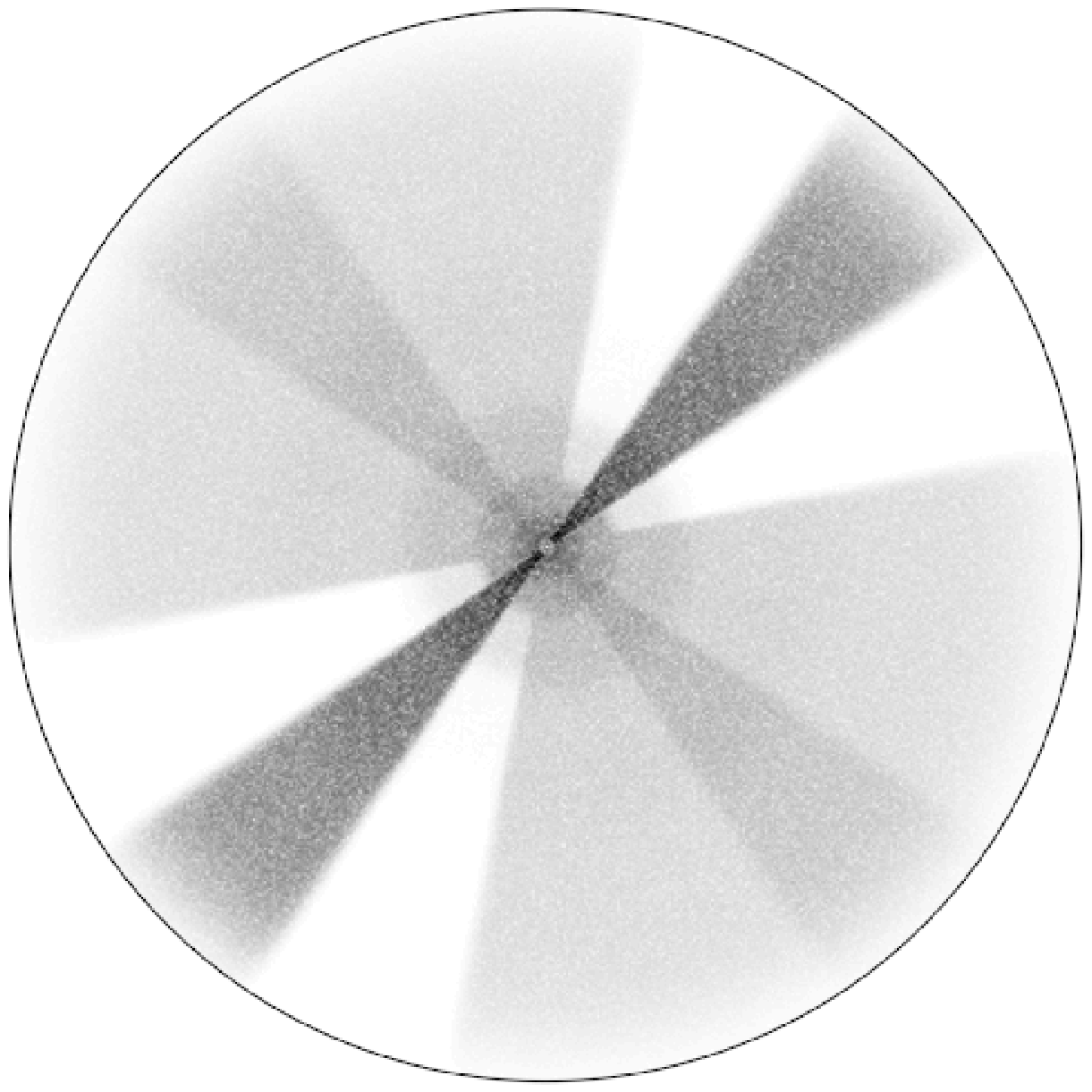}
}
\subfigure[]{
\includegraphics[scale=0.25,trim=0.0in 0 0.0in 0, clip]{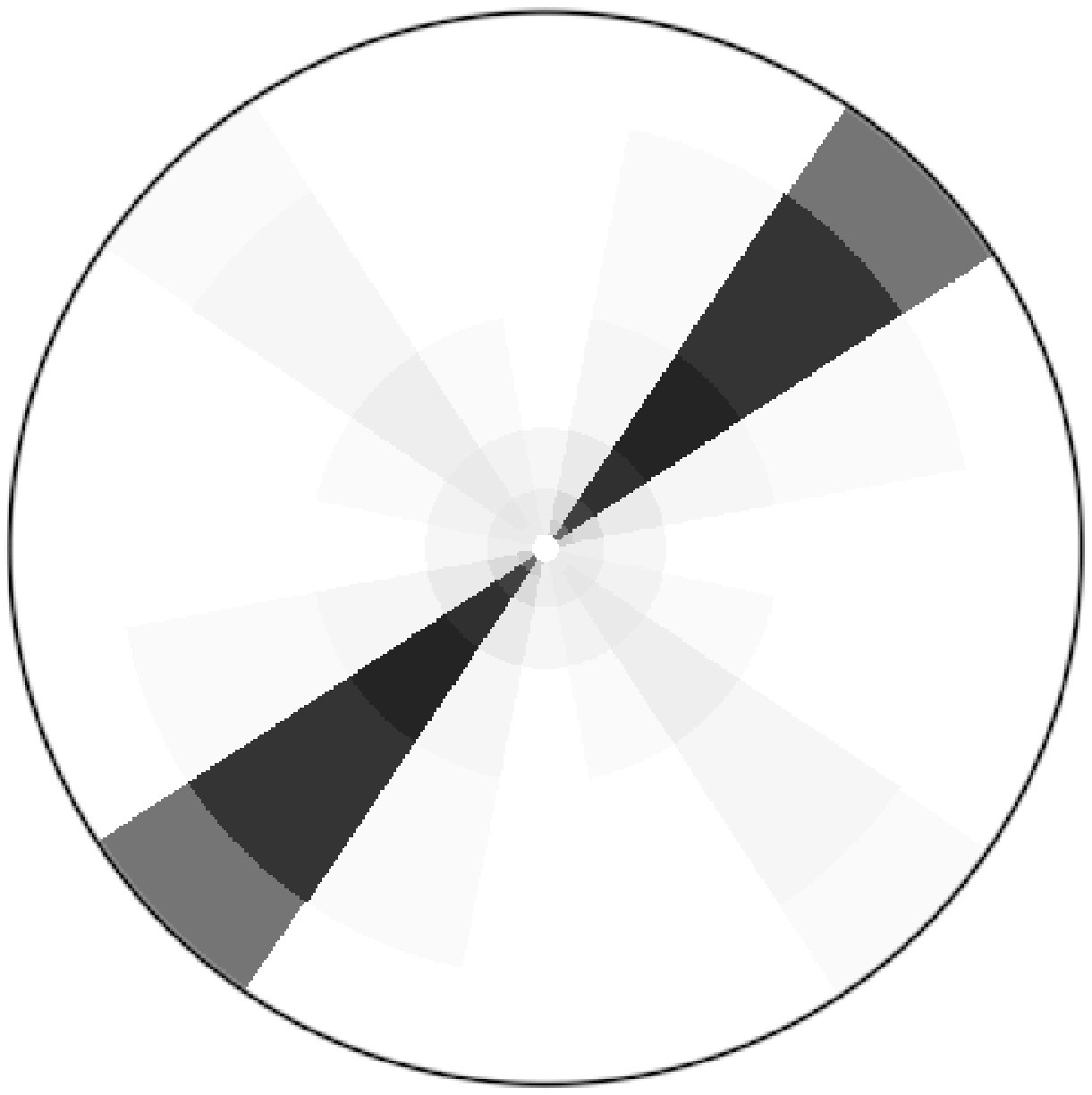}
}
\subfigure[]{
\includegraphics[scale=0.15,trim=0.0in 0 0.0in 0, clip]{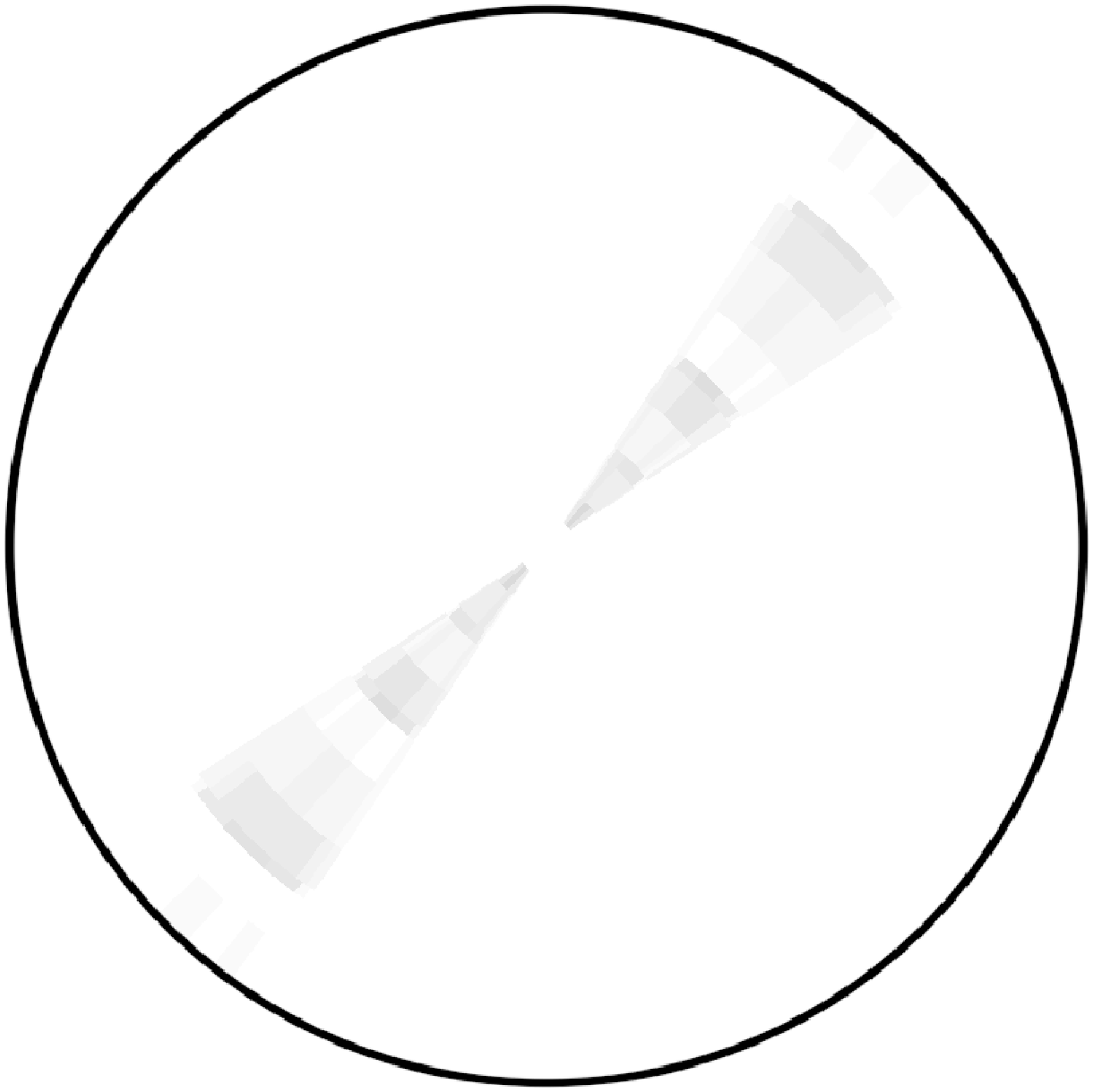}
}
\label{order2_scatfig}
\caption{Scattering display of two images having the same first
order scattering coefficients.
(a) Image $x(u)$. (b) Fourier modulus $|\hat{x}(\om)|$. (c) 
Scattering $S_J x[p(\om)]$ for $m=1$. 
(d) Scattering $S_Jx[p(\om)]$ for $m=2$.
}
\label{order1_order2_comp}
\end{figure*}

Figure \ref{order1_order2_comp} shows the Fourier transform of two images,
and the amplitude of their scattering coefficients of
orders $m=1$ and $m=2$, at a maximum scale $2^J$ equal to the image size.
A scattering coefficient 
over a quadrant $\Omega[2^{j_1} r_1]$
gives an approximation of the
Fourier transform energy over the support of 
$\hat \psi_{2^{j_1} r_1}$. 
Although the top and bottom images are very different, they have 
same order $m=1$ scattering coefficients. 
Here, first-order coefficients are 
not sufficient to discriminate between two very different images.
However, coefficients of order $m=2$ succeed 
in discriminating between the two images.
The top image has wavelet coefficients which are much more sparse
than the bottom image. 
As a result, Section \ref{enerinvsf} shows that
second-order scattering coefficients have a larger amplitude.
Higher-order coefficients are not displayed because they have a negligible
energy as explained in Section \ref{propscasnf}.

\section{Scattering Properties}
\label{propscasnf}

A convolution network is highly non-linear,
which makes it difficult to understand how the coefficient values
relate to the signal properties.
For a scattering network, Section \ref{enerinvsf} analyzes
the coefficient properties and optimizes the network architecture. 
For texture analysis,
the scattering transform of stationary processes is studied in
Section \ref{stationscat}. The regularity of scattering
coefficients can be exploited to reduce the size of a scattering representation,
by using a cosine transform, as shown in Section \ref{CosineSca}. 
Finally, Section \ref{fastscsec} provides a fast computational algorithm.

\subsection{Energy Conservation and Deformation Stability}
\label{enerinvsf}

A windowed scattering
$S_J$ is computed with a cascade of wavelet modulus operators
$U_J$, and its properties thus depend upon the wavelet transform 
properties.
Conditions are given on wavelets to define a scattering
transform which is contracting and preserves the signal norm.
This analysis shows that $\|S_J [p] x\|$
decreases quickly as the length of $p$ increases,
and is non-negligible only over
a particular subset of frequency-decreasing paths. 
Reducing computations to these 
paths defines a convolution network with much fewer
internal and output coefficients. 

The norm of a sequence of transformed 
signals $R x = \{g_n \}_{n \in \Omega}$ is defined
by $\|R x \|^2 = \sum_{n \in \Omega} \|g_n \|^2$.
If $x$ is real and
there exists $\epsilon > 0$ such that for all $\om \in \R^2$
\begin{equation}
\label{pars}
1 - \epsilon \leq |\hat \phi(\om)|^2 + \frac 1 2
\sum_{j=1}^{\infty} \sum_{r \in G} |\hat \psi (2^{-j} r \omega)|^2 \leq 1~,
\end{equation}
then applying the Plancherel formula proves that
$W_J x = \{x \star \phi_J\,,\,x \star \psi_\la \}_{\la \in \Lambda_J}$
satisfies
\begin{equation}
\label{wavecont}
(1 - \epsilon)\, \|x\|^2 \leq \|W_J x \|^2 \leq \|x\|^2 ~,
\end{equation}
with $\|W_J x \|^2 = \|x \star \phi_J \|^2 + \sum_{\lambda \in \Lambda_J}
\|x \star \psi_\lambda \|^2$.
In the following we suppose that 
$\epsilon < 1$ and hence that the wavelet transform is a contracting
and invertible operator, with a stable inverse. 
If $\epsilon = 0$ then $W_J$ is unitary. 
The Morlet wavelet $\psi$ in Figure \ref{waveletfig}
satisfies (\ref{pars}) with $\epsilon=0.25$, together with
$\phi(u) = C \exp(-|u|^2/(2 \sigma_0^2))$
with $\sigma_0 = 0.7$ and $C$ adjusted so that $\int \phi(u)\,du = 1$.
These functions are used
in all classification applications.
Rotated and dilated cubic spline
wavelets are constructed in \cite{mallat} to satisfy (\ref{pars}) with
$\epsilon = 0$.

The modulus is contracting in the sense that $||a| - |b|| \leq |a-b|$.
Since $U_J = \{x \star \phi_J\,,\,|x \star \psi_\la| \}_{\la \in \Lambda_J}$ is
obtained with a wavelet transform $W_J$ followed by modulus, which are
both contractive, it is also contractive:
\[
\|U_J x - U_J y \| \leq \|x - y\|~.
\]
If $W_J$ is unitary then $U_J$ also preserves the signal
norm $\|U_J x\| = \|x\|$.

Let  $\cP_J = \cup_{m \geq 0} \Lambda_J^m$ be the set of all possible paths
of any length $m \in \N$. The norm of $S_J[\cP_J] x = \{S_J[p]x\}_{p \in \cP_J}$ is
$\|S_J[\cP_J] x \|^2 = \sum_{p\in \cP_J} \|S_J[p] x\|^2$.
Since $S_J$ iteratively applies $U_J$ which is contractive,
it is also contractive:
\[
\|S_J x - S_J y \| \leq \|x - y\|~.
\]

If $W_J$ is unitary, $\epsilon = 0$ in (\ref{wavecont}) and
for appropriate wavelets, it is proved in \cite{mallat} that
\begin{equation}
\label{planchsdnfns0}
\|S_J x \|^2 = \sum_{m=0}^{\infty} \|S_J [\Lambda_J^m] x \|^2 =
\sum_{m=0}^{\infty} \sum_{p \in \Lambda_J^m} \|S_J [p] x \|^2 =
\|x \|^2 ~.
\end{equation}

This result uses the fact that $U_J$ preserves the signal norm
and that
$U_J \,U[\Lambda_J^m] x = \{S_J [\Lambda_J^m] x \, , \, U [\Lambda_J^{m+1}] x \}$.
Proving (\ref{planchsdnfns0}) is thus equivalent to 
prove that
the energy of the last network layer converges to zero when $m_{\max}$ increases
\begin{equation}
\label{planchsdnfns89sd}
\lim_{m_{\max} \rightarrow \infty} \|U [\Lambda_J^{m_{\max}}]  x \|^2 =
\lim_{m_{\max} \rightarrow \infty} \sum_{m=m_{\max}}^{\infty} \|S_J [\Lambda_J^m] x \|^2 = 0~.
\end{equation}
This result is also important for numerical applications because 
it explains why the network
depth can be limited with a negligible loss of signal energy. 

The scattering energy conservation also provides a relation between
the network energy distribution and the
wavelet transform sparsity. For $p = (\la_1,...,\la_m)$, we denote
$p+\la = (\la,\la_1,...,\la_m)$.
Applying (\ref{planchsdnfns0}) 
to $U[\la] x = |x \star \psi_{\la}|$ instead of $x$,
and separating the first term for $m=0$ yields
\begin{equation}
\label{planchsdnfns}
\|S_J[\la] x \|^2 + \sum_{m=1}^{\infty} \sum_{p \in \Lambda_J^m} 
\|S_J [\la+p] x \|^2 = \|x \star \psi_\la \|^2 ~. 
\end{equation}
But $S_J [\la] x = |x \star \psi_\la| \star \phi_{2^J}$ is
a local $\LU$ norm and one
can prove \cite{mallat} that 
$\lim_{J \rightarrow \infty} 2^{2 J} \|S_J [\la] x \|^2 = \|\phi\|^2 \, 
\|x \star \psi_\la\|_1^2$.
The more sparse $x \star \psi_\la(u)$ the smaller
$\|x \star \psi_\la\|_1^2$ and 
(\ref{planchsdnfns}) implies that the total
energy $\sum_{m=1}^{\infty} \sum_{p \in \Lambda_J^m} \|S_J [p+\la] x \|^2$ of
higher-order scattering terms is then larger.
Figure \ref{order1_order2_comp} shows two images having same first order scattering 
coefficients, but the top image is piecewise regular and hence has
wavelet coefficients which are much more sparse than the uniform
texture at the bottom. As a result the top image has second order scattering 
coefficients of larger amplitude than at the bottom.
For typical images, as in the CalTech101 dataset \cite{caltech},
Table \ref{scat_energy_absortion1} shows that
the scattering energy has an exponential decay as a function of 
the path length $m$. As proved by
(\ref{planchsdnfns89sd}), the energy of scattering coefficients
converges to $0$ as $m$ increases and is below $1\%$ for $m \geq 3$.

\begin{table}[t]
\caption{This table gives the percentage of scattering energy
$\|S_J(\Lambda_{J}^m)x\|^2/\|x\|^2$ 
captured by frequency-decreasing paths
of length $m$, as a function of $J$. These are 
averaged values computed over 
normalized images with 
$\int x(u)du=0$ and $\|x\|=1$, in the Caltech-101 database. 
The scattering is computed with cubic spline wavelets.}
\label{scat_energy_absortion1}
\begin{center}
\begin{tabular}{|c | c c c c c |c|}
\hline
$J$ & $m=0$ & $m=1$ & $m=2$ & $m=3$ & $m=4$ & $m \leq 3$\\ 
\hline
$1$ & 95.1 & 4.86 & - & - & - & 99.96\\ 
$2$ & 87.56 & 11.97 & 0.35 & - & -& 99.89\\ 
$3$ & 76.29 & 21.92 & 1.54 & 0.02 & -& 99.78\\ 
$4$ & 61.52 & 33.87 & 4.05 & 0.16 & 0& 99.61\\ 
$5$ & 44.6 & 45.26 & 8.9 & 0.61 & 0.01&99.37\\ 
$6$ & 26.15 & 57.02 & 14.4 & 1.54 & 0.07& 99.1\\ 
$7$ & 0 & 73.37 & 21.98 & 3.56 & 0.25& 98.91\\ 
\hline
\end{tabular}
\end{center}
\end{table}

The energy conservation (\ref{planchsdnfns0}) is proved by showing
that the scattering
energy $\|U[p] x \|^2$ 
propagates towards lower frequencies as the length of $p$ increases.
This energy is thus ultimately captured by the low-pass filter 
$\phi_{2^J}$ which outputs $S_J [p] x = U[p] x \star \phi_{2^J}$. 
This property requires that $x \star \psi_\la$ has a lower-frequency
envelope $|x \star \psi_\la|$. It is valid if
$\psi(u) = e^{i \eta.u}\, \theta(u)$ where $\theta$
is a low-pass filter. To verify this property, we write
$x \star \psi_\la (u) = e^{i \la \xi.u}\, x_\la (u)$ with
\[
x_\la(u) = (e^{-i \la \xi.u} x(u)) \star \theta_\la (u)~.
\]
This signal is filtered by the dilated and rotated
low-pass filter $\theta_\la$ whose Fourier transform is
$\hat \theta_\la (\om) = \theta(\la^{-1} \om)$. 
So $|x \star \psi_\la (u)| = |x_\la (u)|$ 
is the modulus of a regular function
and is therefore mostly regular.
This result is not valid if $\psi$ is a real
because $|x \star \psi_\la|$ is singular at each zero-crossing of
$x \star \psi_\la(u)$.

The modulus appears as a non-linear  ``demodulator'' which
projects wavelet coefficients to lower frequencies. 
If $\la = 2^j r$ then
$|x \star \psi_\la (u)| \star \psi_{\la'}$ for  $\la' = 2^{j'} r'$ is
non-negligible only if $\psi_{\la'}$ is located at low frequencies
and hence if $2^{j'} < 2^j$. 
Iterating on wavelet modulus operators thus propagates the scattering
energy along frequency-decreasing paths
$p = (2^{j_1} r_1,...,2^{j_m} r_m)$ where $2^{j_k} \leq 2^{j_{k-1}}$,
for $1\leq k < m $. Scattering coefficients along other paths have
a negligible energy.
Over the CalTech101 images database, 
Table \ref{scat_energy_absortion1} shows that 
over $99\%$ of the scattering energy is concentrated
along frequency-decreasing paths of length $m \leq 3$.
Numerically, it is
therefore sufficient to compute the scattering transform along this
subset of frequency-decreasing paths. It defines a much smaller convolution
network. Section \ref{fastscsec} shows that the resulting
coefficients are computed with $O(N \log N)$ operations.

For classification applications, one of the most important properties
of a scattering transform is its stability to deformations
$L_\tau x(u) = x(u-\tau(u))$, 
because wavelets are stable to deformations
and the modulus commutes with $L_\tau$. 
Let $\|\tau \|_\infty = \sup_u |\tau (u)|$ and
$\|\nabla \tau \|_\infty = \sup_u |\nabla \tau (u)| < 1$.
If $S_J$ is computed on paths of length $m \leq m_{\max}$
then it is proved in
\cite{mallat} that for signals $x$ of compact support
\begin{equation}\label{lipschitz}
\|S_{J} (L_\tau x) - S_{J}x\| \leq C m_{\max} \|x\| 
\Big(2^{-J} \|\tau\|_\infty +  \|\nabla \tau \|_\infty \Big)~,
\end{equation}
with a second order Hessian term which is negligible if $\tau(u)$ is regular.  
If $2^J \geq \|\tau \|_\infty / \|\nabla \tau \|_\infty$ then 
the translation term can be neglected and the transform 
is Lipschitz continuous to deformations:
\begin{equation}\label{lipschitz2}
\|S_{J} (L_\tau x) - S_{J}x\| \leq C m_{\max} \|x\| \|\nabla \tau\|_\infty  ~.
\end{equation}

\subsection{Scattering Stationary Processes}
\label{stationscat}

\begin{figure*}[ht]
\centering
\subfigure{
\includegraphics[scale=0.3,trim=1.3in 0 1.1in 0, clip]{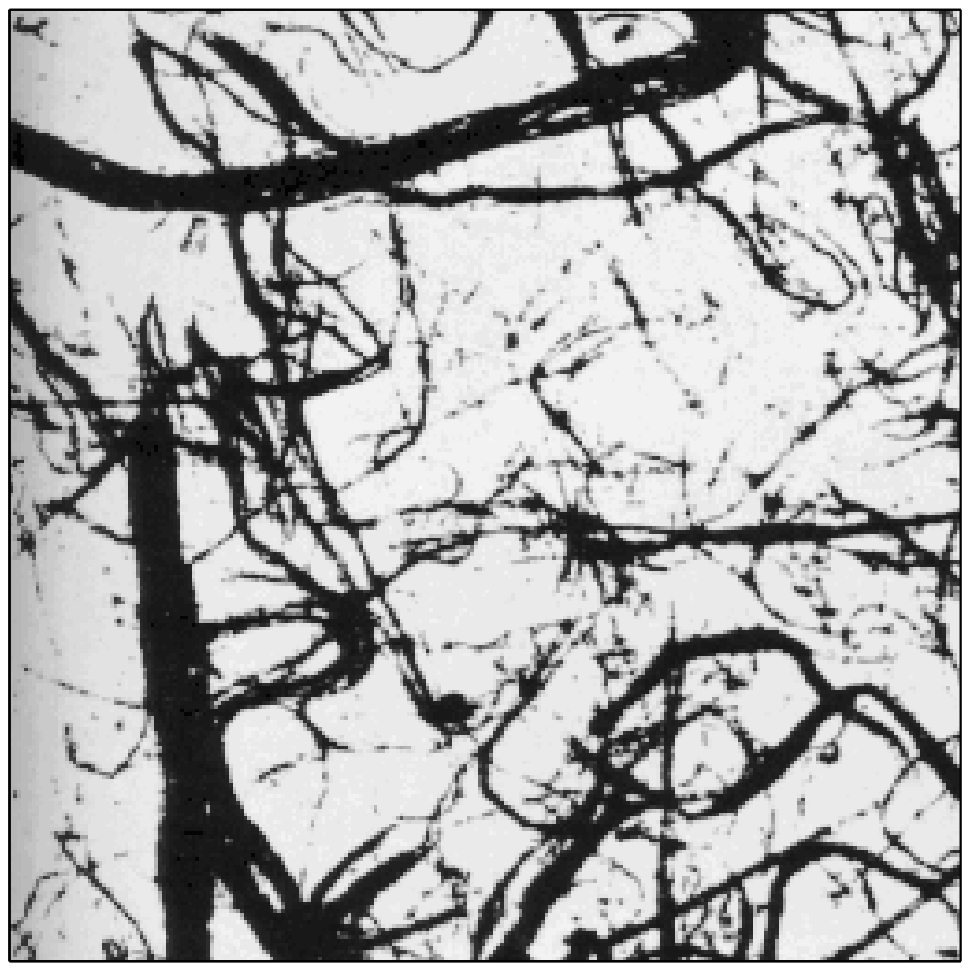}
}
\subfigure{
\includegraphics[scale=0.3,trim=1.3in 0 1.1in 0, clip]{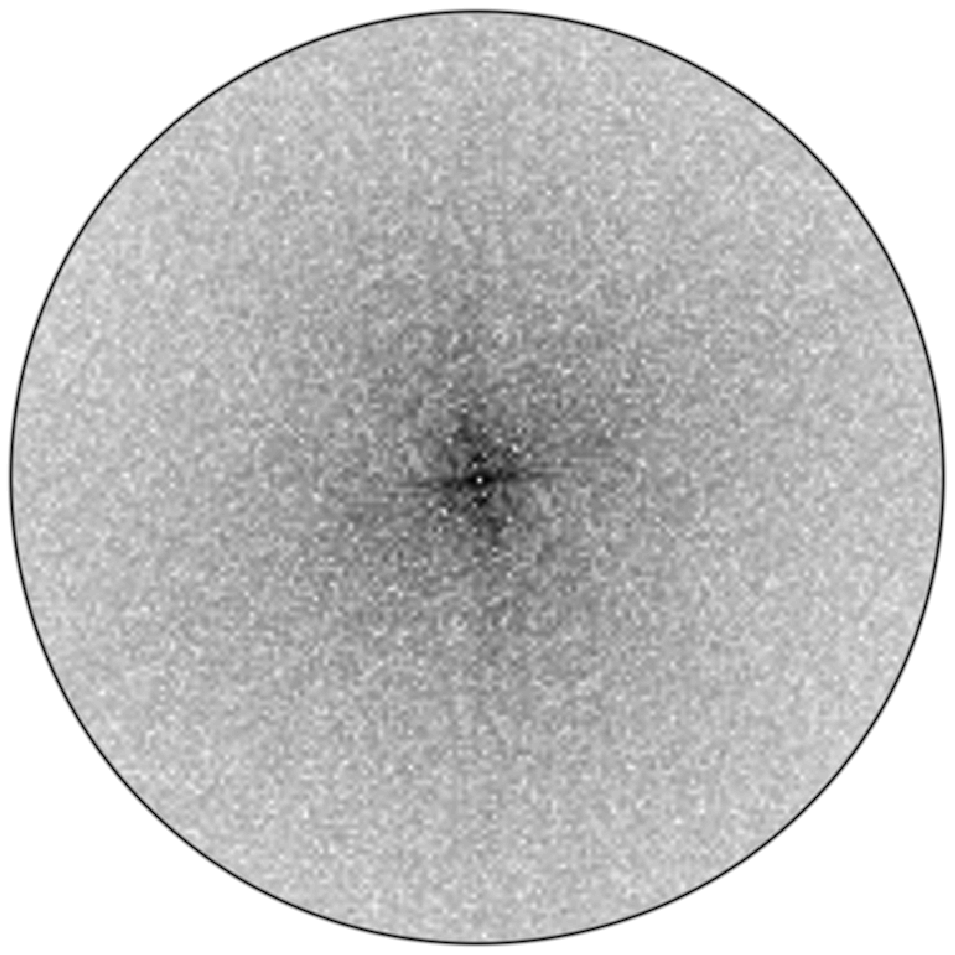}
}
\subfigure{
\includegraphics[scale=0.3,trim=1.3in 0 1.1in 0, clip]{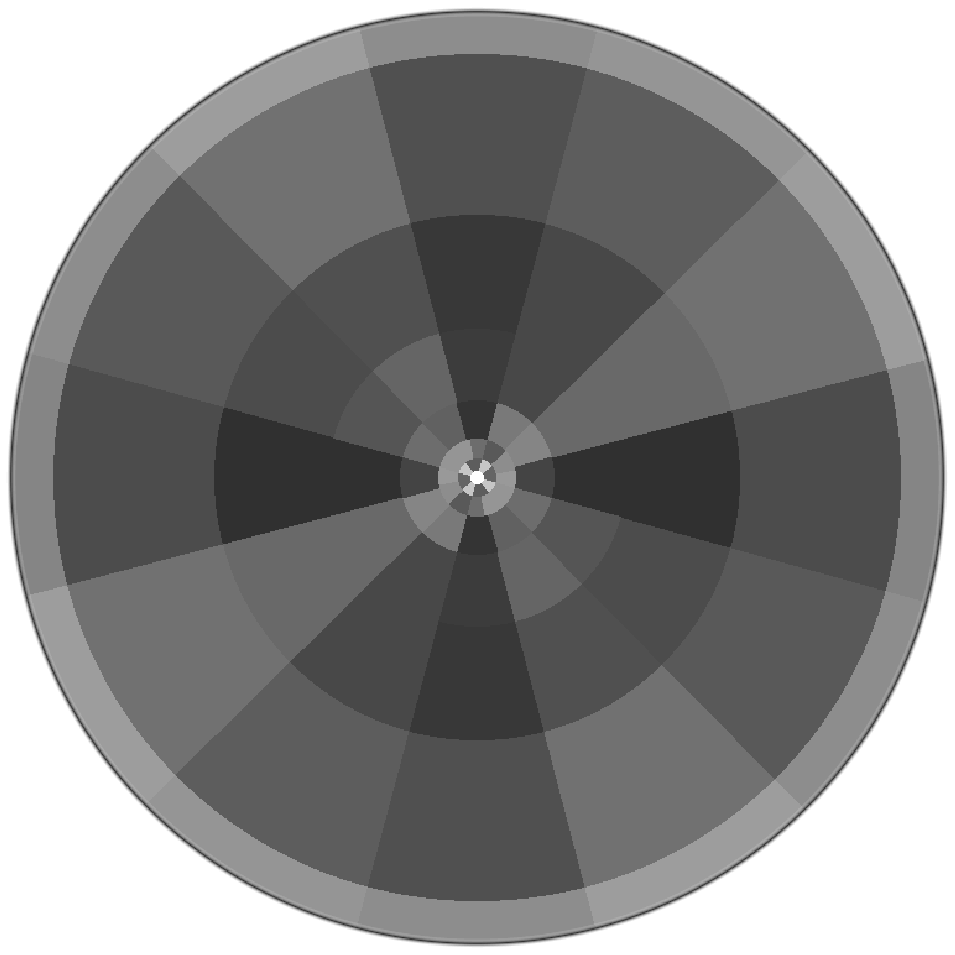}
}
\subfigure{
\includegraphics[scale=0.3,trim=1.3in 0 1.1in 0, clip]{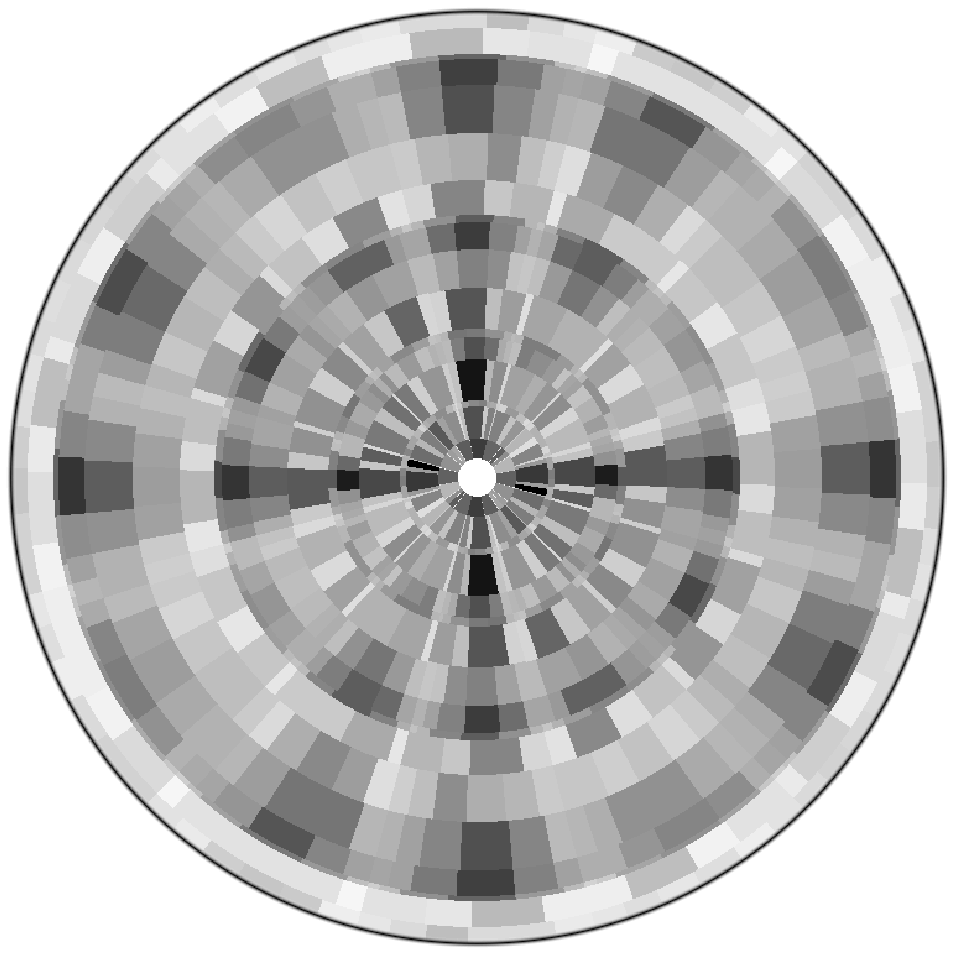}
} \\
\setcounter{subfigure}{0}
\subfigure[]{
\includegraphics[scale=0.3,trim=1.3in 0 1.1in 0, clip]{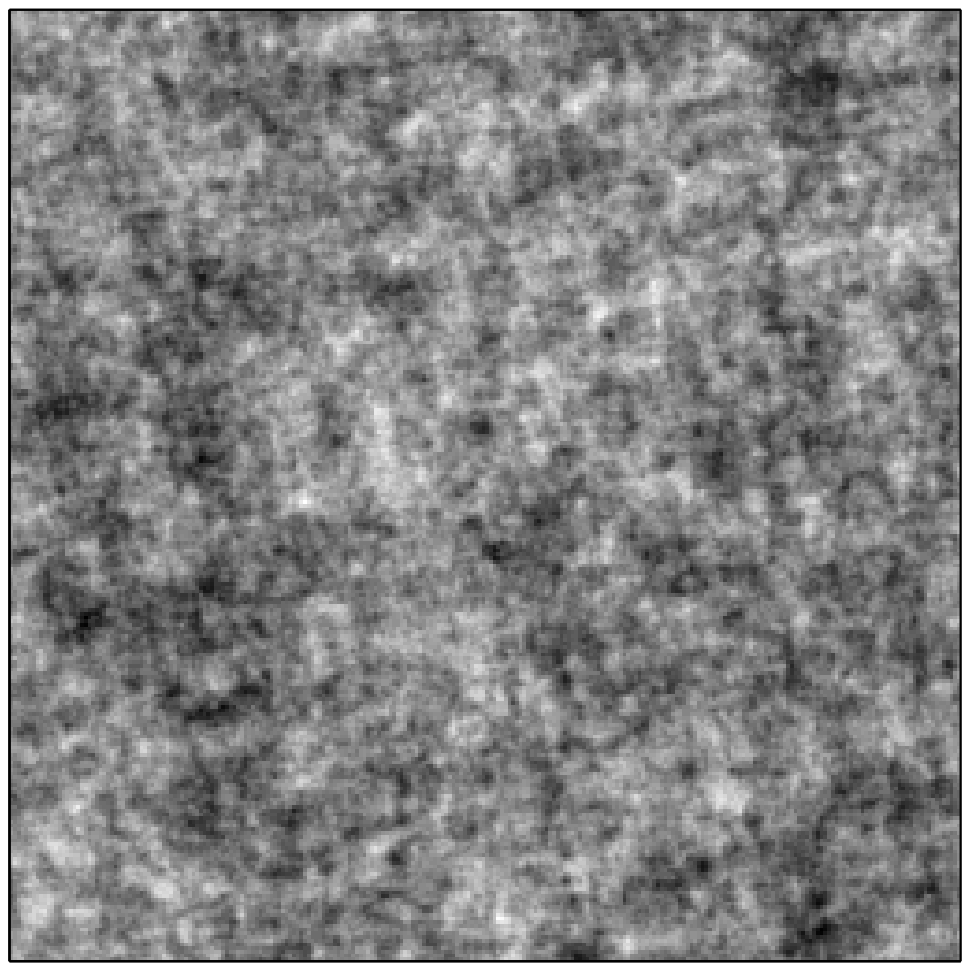}
}
\subfigure[]{
\includegraphics[scale=0.3,trim=1.3in 0 1.1in 0, clip]{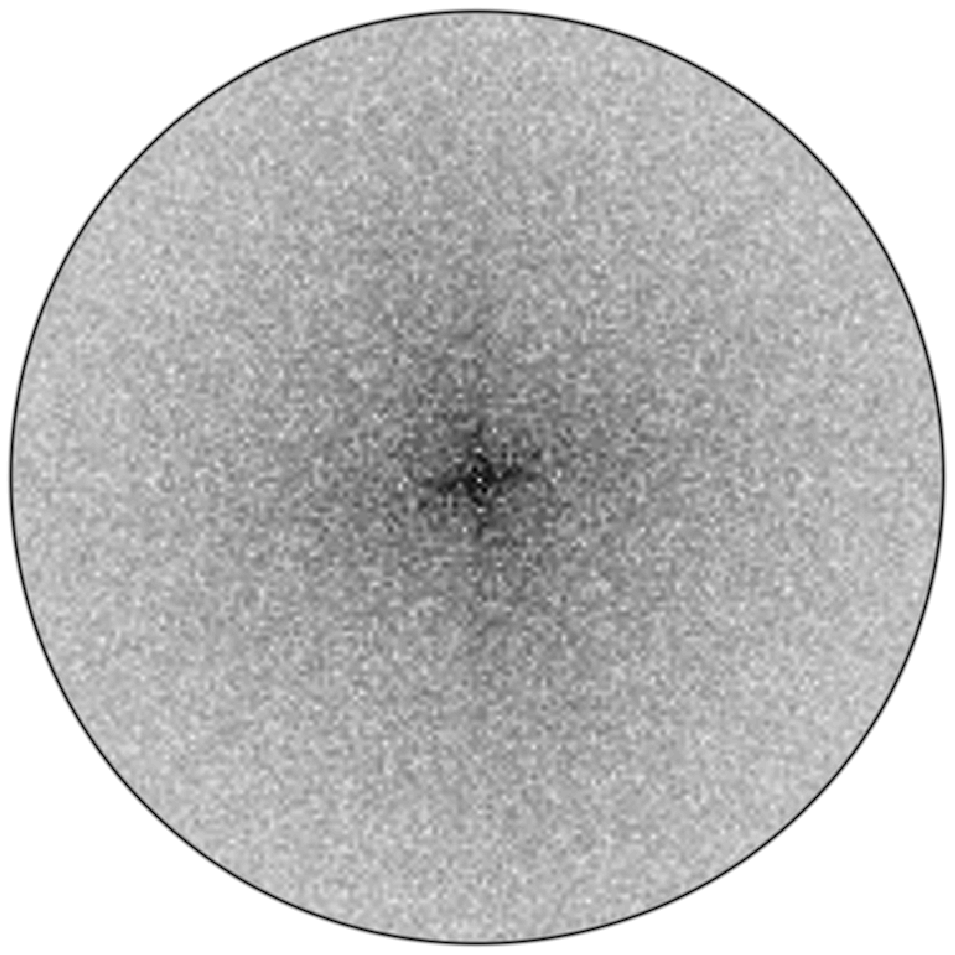}
}
\subfigure[]{
\includegraphics[scale=0.3,trim=1.3in 0 1.1in 0, clip]{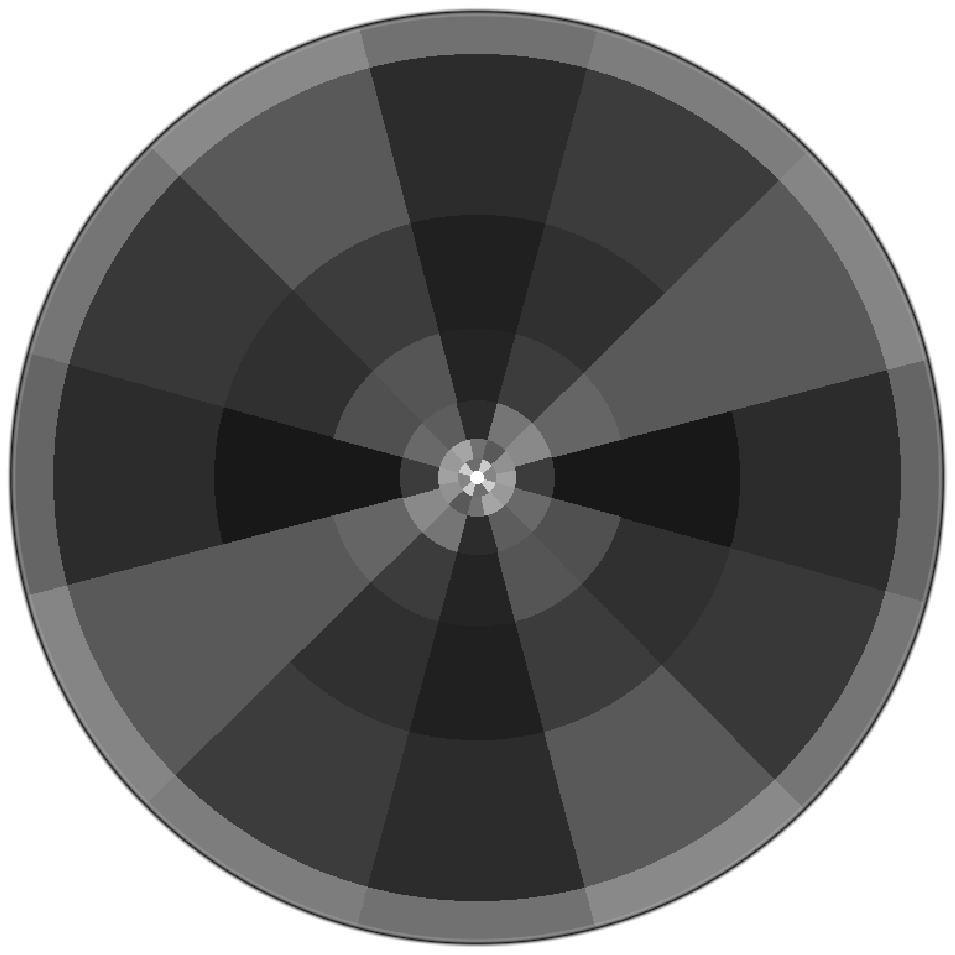}
}
\subfigure[]{
\includegraphics[scale=0.3,trim=1.3in 0 1.1in 0, clip]{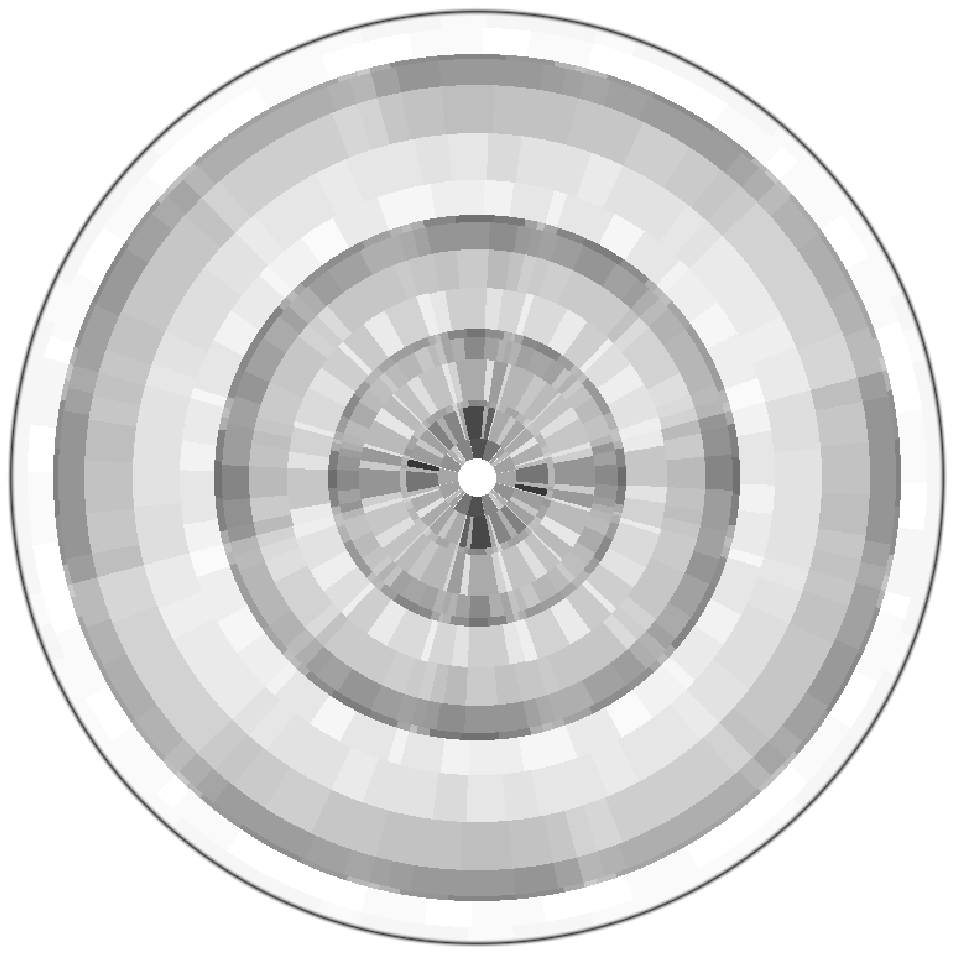}
}
\caption{Two different textures having the same 
Fourier power spectrum. 
(a) Textures $X(u)$. Top: Brodatz texture. Bottom: 
Gaussian process.
(b) Same estimated power spectrum $\widehat R X(\omega)$.
(c) Nearly same scattering coefficients $S_J [p] X$ for $m=1$ 
and $2^J$ equal to the image width.
(d) Different scattering coefficients $S_J[p] X$ for $m=2$.
}
\label{norm_textures}
\end{figure*}

Image textures can be modeled as realizations
of stationary processes $X(u)$. We denote the expected value
of $X$ by $E(X)$, 
which does not depend upon $u$. 
The Fourier spectrum $\widehat RX (\omega)$ 
is the Fourier transform of the autocorrelation
\[
RX (\tau) = E \Big([X(u) - E (X)] [X(u-\tau) - E (X)] \Bigr)~.
\]
Despite 
the importance of spectral methods,
the Fourier spectrum is often
not sufficient to discriminate image textures because it does not take
into account higher-order moments.
Figure \ref{norm_textures} shows very different textures having
same second-order moments.
A scattering representation of stationary processes includes second order and
higher-order moment descriptors
of stationary processes, which discriminates between such textures.

\setcounter{subfigure}{0}

If $X(u)$ is stationary then $U[p] X(u)$ remains stationary because
it is computed with  a cascade of convolutions
and modulus operators which preserve stationarity. Its expected value
thus does not depend upon $u$ and defines the expected scattering transform:
\[
\overline S X(p) = E (U[p] X)~.
\]

A windowed scattering gives an estimator of 
$\overline S X(p)$, calculated from a single realization of $X$,
by averaging $U[p] X$ with $\phi_{2^J}$:
\[
S_J [p] X(u) = U[p] X \star \phi_{2^J} (u)~.
\]
Since $\int \phi_{2^J} (u)\,du = 1$, this estimator is unbiased:
$E (S_J[p] X) = E (U[p] X)$.

For appropriate wavelets, it is
also proved \cite{mallat} that 
\begin{equation}
\label{varndosdfn98sd}
\sum_{p \in \cP_J} E (|S_J[p]X|^2 ) = E(|X|^2)~.
\end{equation}
Replacing $X$ by $X \star \psi_\la$ implies that
\[
\sum_{p \in \cP_J} E (|S_J[p+\la]X|^2 ) = E(|X \star \psi_\la|^2)~.
\]
These expected squared wavelet coefficients can also be written
as a filtered integration of the Fourier power spectrum $\widehat RX(\om)$
\[
E(|X \star \psi_\la|^2)
= \int \widehat RX (\om) \, |\hat \psi(\lambda^{-1} \om)|^2\, d \om~.
\]
These two equations prove that 
summing scattering coefficients recovers the 
power spectrum integral over each wavelet frequency support,
which only depends upon second-order moments.
However, one can also show that scattering coefficients 
$\overline S X(p)$ depend upon moments 
of $X$ up to the order $2^m$ if $p$ has a length $m$. 
Scattering coefficients can thus discriminate
textures having same second-order moments but different higher-order moments.
This is illustrated using the two textures in Figure \ref{norm_textures},
which have the same power spectrum and hence same second order moments.
Scattering coefficients $S_J[p] X$ are shown for $m=1$ and $m=2$
with the frequency tiling illustrated in Figure \ref{scat_display_construction}.
The ability to discriminate  
 the top process $X_1$ from the 
bottom process $X_2$
is measured by a scattering distance normalized by the variance:
$$\rho(m) = \frac{\| S_J X_1 [\Lambda_J^m] - E(S_J X_2 [\Lambda_J^m])\|^2}
{E(\| S_J X_2 [\Lambda_J^m] - E(  S_J X_2 [\Lambda_J^m]) \|^2)} ~.$$
For $m=1$, scattering coefficients mostly depend upon second-order
moments and are thus nearly equal for both textures. One can indeed
verify numerically that $\rho(1) = 1$ so both textures can not be distinguished
using first order scattering coefficients.
On the contrary, scattering coefficients of order $2$ 
are highly dissimilar because they depend on moments up to order $4$,
and $\rho(2)=5$.

For a large class of ergodic processes including
most image textures, it is observed numerically that the total scattering
variance
$\sum_{p \in \cP_J} E (|S_J [p] X - \overline SX(p)|^2 )$ decreases to zero
when $2^J$ increases. Table \ref{brodatz_variance_decay} 
shows the decay of the total scattering
variance, computed on average over the Brodatz texture dataset.
Since $E(|S_J[p]X|^2) = E (S_J [p] X)^2 + E( |S_J [p] X-E( S_J [p] X)|^2)$
and $E (S_J [p] X) = \overline SX(p)$, 
it results from the energy
conservation (\ref{varndosdfn98sd}) 
that the expected scattering transform also satisfies 
\[
\|\overline S X \|^2 = \sum_{m=0}^\infty \sum_{p \in \Lambda_\infty^m}
|\overline SX(p)|^2 = E (|X|^2)~.
\]
Table \ref{scat_brodatz_order} gives the percentage of expected
scattering energy
$\sum_{p \in \Lambda_\infty^m}|\overline SX(p)|^2$ carried by paths of length $m$,
for textures in the Brodatz
database. Most of the energy is concentrated in paths of length
$m \leq 3$. 

\begin{table}[t]
\caption{Decay of the total scattering variance 
$\sum_{p \in \cP_J} E (|S_J [p] X - \overline SX(p)|^2 ) / E(|X|^2)$ 
in percentage, as a function of $J$, averaged over the Brodatz dataset. 
Results obtained using cubic spline wavelets. }
\label{brodatz_variance_decay}
\begin{center}
\begin{tabular}{| c c c c c c c |}
\hline
$J=1$ & $J=2$ & $J=3$ & $J=4$ & $J=5$ & $J=6$ & $J=7$\\ 
\hline
85 & 65 & 45 & 26 & 14 & 7 & 2.5\\ 
\hline
\end{tabular}
\end{center}
\end{table}

\begin{table}[t]
\caption{Percentage of expected scattering energy 
$\sum_{p \in \Lambda_\infty^m}|\overline SX(p)|^2$,
as a function of the scattering order $m$, computed with cubic spline wavelets,
over the Brodatz dataset.}
\label{scat_brodatz_order}
\begin{center}
\begin{tabular}{| c c c c c |}
\hline
$m=0$ & $m=1$ & $m=2$ & $m=3$ & $m=4$\\ 
\hline
0 & 74 & 19 & 3 & 0.3\\ 
\hline
\end{tabular}
\end{center}
\end{table}


\subsection{Cosine Scattering Transform}
\label{CosineSca}

\begin{figure*}[ht]
\centering
\subfigure{
\includegraphics[scale=0.29,trim=0.8in 0 0.8in 0, clip]{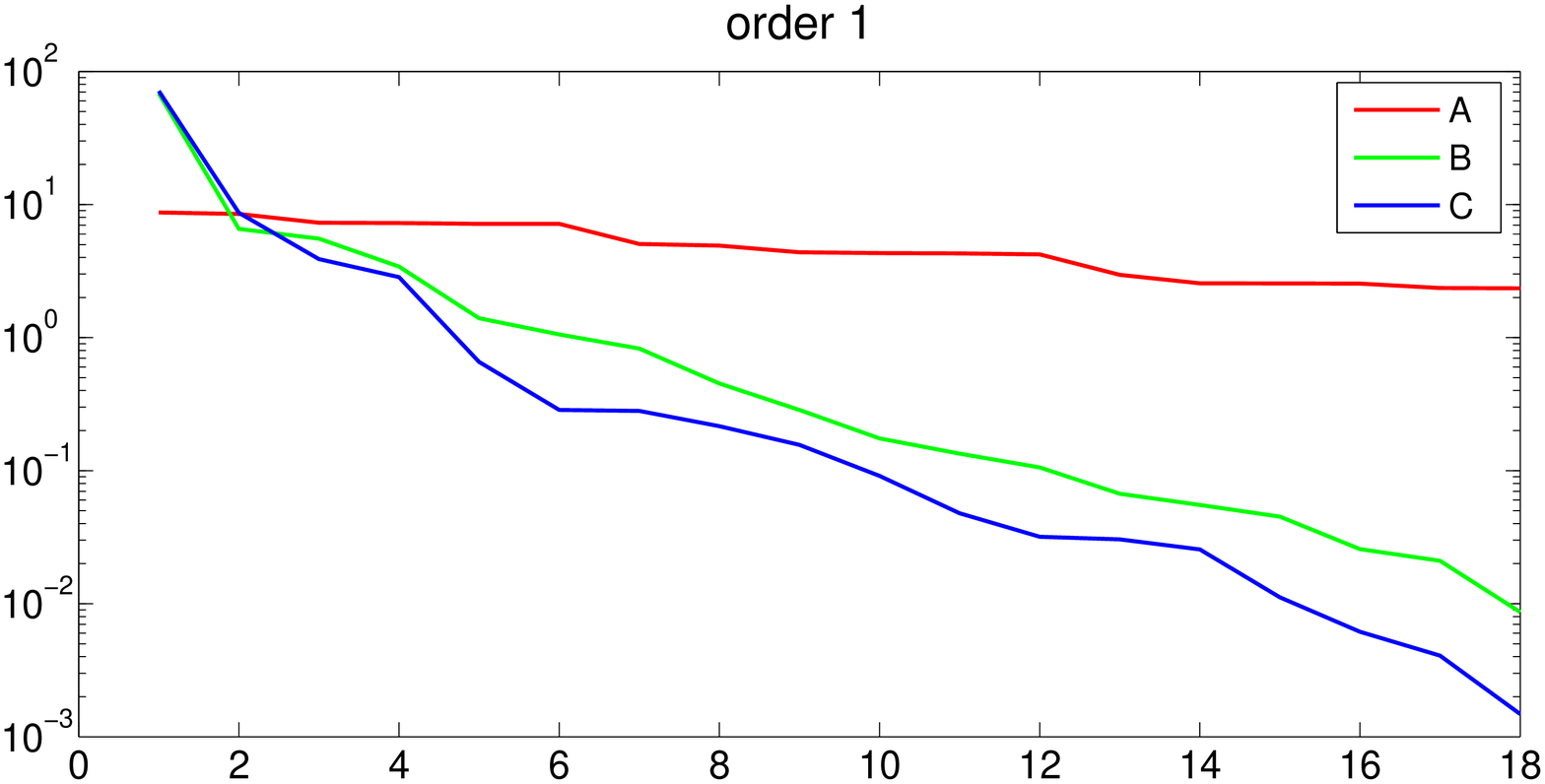}
}
\subfigure{
\includegraphics[scale=0.29,trim=0.8in 0 0.8in 0, clip]{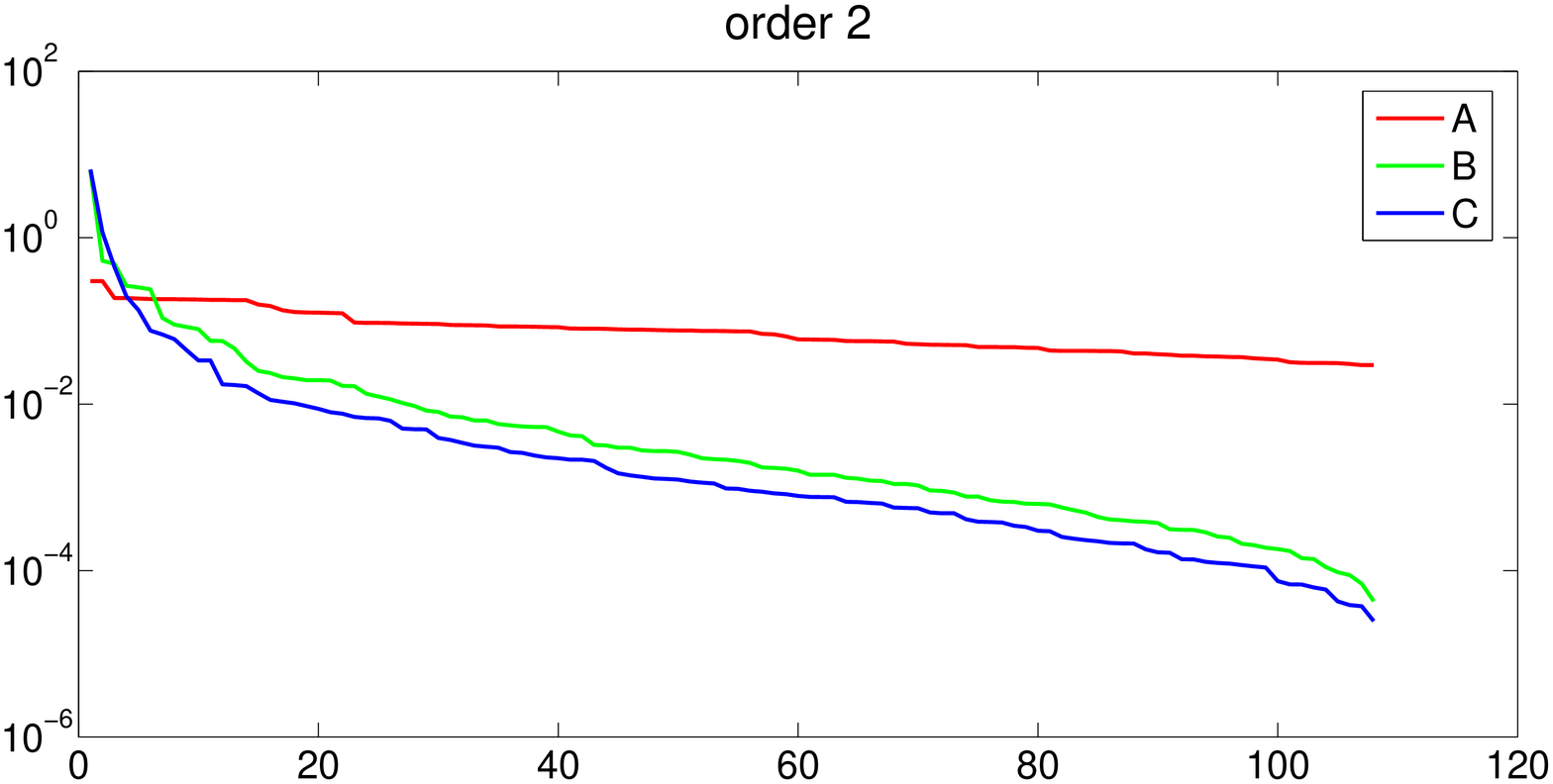}
}
\caption{(A): Sorted variances of scattering coefficients for $m=1$ (left)
and $m=2$ (right).
(B): Sorted variances of DCT scattering coefficients. 
(C): Variances in the scattering Karhunen-Loeve basis.}
\label{dct_figures}
\end{figure*}

Natural images have scattering coefficients $S_J [p] X(u)$ 
which are correlated across paths $p = (2^{j_1} r_1,...,2^{j_m} r_m)$, 
at any given position $u$. The strongest correlation is between paths
of same length. For each $m$,
scattering coefficients are decorrelated in 
a Karhunen-Lo\`eve basis which
diagonalizes their covariance matrix. 
Figure \ref{dct_figures} compares the decay of the sorted
variances $E (|S_J [p] X - E(S_J[p] X)|^2 )$ and the variance decay
in the Karhunen-Loève basis computed on paths of length
$m=1$, and on paths of length $m=2$, 
over the Caltech image dataset with a Morlet wavelet.
The variance decay is much faster in the Karhunen-Loève basis, 
which shows 
that there is a strong correlation between scattering coefficients
of same path length.

A change of variables proves 
that a rotation and scaling $X_{2^l r} (u) = X(2^{-l} r u)$ produces
a rotation and inverse scaling on the path variable
$p = (2^{j_1} r_1,...,2^{j_m} r_m)$:
\[
\overline S X_{2^l r} (p) = \overline S X(2^l r p)~~\mbox{where}~~
2^l r p = (2^{l+j_1} r r_1,...,2^{l+j_m} r r_m)~.
\]
If images are randomly rotated and scaled
by $2^l r^{-1}$ then the path $p$ 
is randomly rotated and scaled \cite{perrinet}. 
In this case, the scattering transform has stationary variations
along the scale and rotation variables. This suggests approximating the
Karhunen-Loève basis by a cosine basis along these variables.
Let us parameterize each rotation
$r$ by its angle $\theta \in [0,2\pi]$.
A path $p$ is then parameterized
by $([j_1,\theta_1],...,[j_m,\theta_m])$. 

Since scattering coefficients are computed along frequency
decreasing paths for which $-J \leq j_k < j_{k-1}$,
to reduce boundary effects,
a separable cosine transform is computed along the 
variables
$\tilde{j}_1=j_1\,,\,\tilde{j}_2=j_2-j_1,\, ...\,, \tilde{j}_m=j_m-j_{m-1}$,
and along each angle variable $\theta_1,\,\theta_2,\,...\,,\theta_m$.
We define the cosine scattering transform as the coefficients obtained
by applying this separable
discrete cosine transform along the scale and angle variables of
$S_J[p] X(u)$, for each $u$ and each path length $m$.
Figure \ref{dct_figures} shows that the cosine scattering coefficients have 
variances for $m=1$ and $m=2$ which decay nearly as fast as the
variances in the Karhunen-Loeve basis. It shows that
a DCT across scales and orientations
is nearly optimal to decorrelate scattering coefficients.
Lower-frequency DCT coefficients absorb most of the scattering energy.
On natural images, more than
99\% of the scattering energy is absorbed by the $1/3$ lowest frequency
cosine scattering coefficients.

\setcounter{subfigure}{0}

\subsection{Fast Scattering Computations}
\label{fastscsec}

Section \ref{enerinvsf} shows that the scattering energy
is concentrated 
along frequency-decreasing paths $p = (2^{j_k} r_k)_k$ satisfying
$2^{-J} \leq 2^{j_{k+1}} < 2^{j_k}$. 
If the wavelet transform is computed along $C$ directions then 
the total number of frequency-decreasing paths of length $m$
is $C^m\binom{J}{m}$. 
Since $\phi_{2^J}$ is a low-pass filter,
$S_J [p] x(u) = U[p] x \star \phi_{2^J} (u)$ can be uniformly
sampled at intervals $\alpha 2^{J}$, with 
$\alpha = 1$ or $\alpha  = 1/2$.
If $x(n)$ is a discrete image with
$N$ pixels, then each $S_J [p] x$
has $2^{-2J} \alpha^{-2} N$ coefficients. 
The scattering representation along all frequency-decreasing paths
of length at most $m$ thus has a total 
number of coefficients equal
to $N_J = N \alpha^{-2} 2^{-2J} \sum_{q=0}^{m} C^q\binom{J}{q}$.
This reduced scattering representation is computed by 
a cascade of convolutions, modulus, and sub-samplings, with
$O(N \log N)$ operations. The final DCT transform further compresses
the resulting representation.

Let us recall from Section \ref{convnetwosec} that
scattering coefficients are computed by iteratively
applying the one-step propagator $U_J$. To compute subsampled
scattering coefficients along frequency-decreasing paths, 
this propagator is truncated. For any scale $2^k$,
${U}_{k,J}$ transforms a signal $x(2^k \alpha n)$ into
\begin{equation}
\label{wavedfn}
{U}_{k,J} x = \Big\{
x \star \phi_J (2^J \alpha n)~,~
|x \star \psi_{2^j r} (2^j \alpha n)| \Big\}_{-J < j \leq k , r \in G^+}~.
\end{equation}
The algorithm computes subsampled scattering coefficients
by iterating on this propagator.

\begin{algorithm}\caption{Reduced Scattering Transform}\label{scatt-algo}
\begin{algorithmic}
\STATE Compute $U_{0,J} (x)$
\STATE Output $x \star \phi_{2^J} (2^J \alpha n)$
\FOR{$m =1$ to $m_{\max}-1$}
\FORALL{$0\geq j_1 > ...>j_m > -J$}
\FORALL{$(r_1,...,r_q) \in G^{+m}$}
\IF{$m = m_{\max}-1$}
\STATE Compute 
$|||x \star \psi_{2^{j_1}r_1}| \star ... |\star \psi_{2^{j_m}r_m}| \star \phi_{2^J} 
(2^J \alpha n)$
\ELSE
\STATE Compute 
${U}_{j_m,J} (|||x \star \psi_{2^{j_1}r_1}| \star ... |\star \psi_{2^{j_m}r_m}|) $
\ENDIF
\STATE Output $|||x \star \psi_{j_1,\ga_1}| \star ... |\star \psi_{j_q,\ga_q}| 
\star \phi_{J} (2^J \alpha n)$
\ENDFOR
\ENDFOR
\ENDFOR
\end{algorithmic}
\end{algorithm}

If $x$ is a signal of size $P$ then FFT's compute $U_{k,J} x$ 
with $O(P \log P)$ operations.
A reduced scattering transform thus 
computes its $N_J = N \alpha^{-2} 2^{-2J} \sum_{m=0}^{m_{\max}} C^m\binom{J}{m}$ 
coefficients with $O(N_J \log N)$ operations. If $m_{\max} = 2$
then $N_J = N \alpha^{-2} 2^{-2J} (C J + C^2 J(J-1)/2)$. It decreases
exponentially when the scale $2^J$ increases.

Scattering coefficients are decorrelated
with a separable DCT along each scale variable 
$\tilde{j}_1=j_1\,,\,\tilde{j}_2=j_2-j_1,\, ...\,, \tilde{j}_m=j_m-j_{m-1}$
and each rotation
angle variable $\theta_1,\,\theta_2,\,...\,,\theta_m$, which
also requires $O(N_J \log N)$ operations. 
For natural images, more than 99.5 \% of 
the total signal energy is carried by the resulting $N_J / 2$ cosine scattering 
coefficients of lower frequencies.

Numerical computations in this paper are performed  
by rotating wavelets along $C = 6$ directions, for scattering
representations of maximum order $m_{\max} = 2$.
The resulting size of a reduced cosine scattering representation 
has at most three times as many coefficients as a dense SIFT representation.
SIFT represents small blocks of $4^2$ pixels with $8$ coefficients.
A cosine scattering representation
represents each image block of $2^{2J}$ pixels by 
$N_J 2^{2J}/(2 N) = (C J + C^2 J(J-1)/2)/2$ coefficients, which is equal
to $24$ for $C=6$ and $J=2$. 
The cosine scattering transform is thus three times the size of SIFT for $J=2$, but
as $J$ increases, the relative size decreases.
If $J=3$ then the size of a cosine scattering representation
is twice the size of a SIFT representation
but for $J=7$ it is about $20$ times \emph{smaller}.

\section{Classification Using Scattering Vectors}
\label{numeric}

A scattering transform eliminates the image variability due to translation
and is stable to deformations. The resulting classification properties are
studied with a PCA and an SVM classifier applied to scattering representations
computed with a Morlet wavelet.
State-of-the-art results are obtained for
hand-written digit recognition and for texture discrimination.

\subsection{PCA Affine Scattering Space Selection}
\label{secaffi}

Although discriminant classifiers such as SVM have better asymptotic
properties than generative classifiers \cite{gen_vs_disc}, the situation can
be inverted for small training sets. 
We introduce a simple
robust generative classifier based on affine space models
computed with a PCA. Applying a DCT on scattering coefficients has no
effect on any linear classifier because it is 
a linear orthogonal transform.
However, keeping the $50$\% lower frequency cosine scattering coefficients reduces
computations and has a negligible effect on classification results. 
The classification algorithm is described directly 
on scattering coefficients to simplify explanations.
Each signal class is represented by 
a random vector $X_k$, whose realizations are images of $N$ pixels
in the class.

Let $E (S_J X) = \{E ({S}_J[p] X(u)) \}_{p,u}$ be 
the family of $N_J$ expected scattering values, computed along all 
frequency-decreasing paths of length $m \leq m_{\max}$ and all subsampled positions 
$u = \alpha 2^J n$.
The difference ${S}_J X_k - E({S}_J X_k)$ is approximated by its projection
in a linear space of low dimension $d \ll N_J$.
The covariance matrix of ${S}_J X_k$ is a matrix of size $N_J^2$.
Let ${\mathbf{V}_{d,k}}$ be the linear space generated 
by the $d$ PCA eigenvectors of this covariance matrix 
having the largest eigenvalues.
Among all linear spaces of dimension $d$, this is the space
which approximates ${S}_J X_k - E({S}_J X_k)$ with the smallest
expected quadratic error. 
This is equivalent
to approximating $ {S}_J X_k$ by its projection on an affine approximation
space:
\[
{\bf A}_{d,k} = E \{{S}_J X_k\} + {\bf V}_{d,k} .
\]

The resulting classifier associates a signal $X$ to the 
class $\hat k$ which yields the best approximation space:
\begin{equation}
\label{clasdifnsdf}
\hat{k}(X) = \argmin_{k \leq K}
\|{S}_J X - P_{\mathbf{A}_{d,k}}({S}_J X) \|~.
\end{equation}

The minimization of this distance has similarities with the minimization
of a tangential distance \cite{tangent} in the sense that we remove
the principal scattering directions of variabilities to evaluate the distance.
However it is much simpler since it does not evaluate a tangential space
which depends upon ${S}_J x$.
Let $\mathbf{V}_{d,k}^\perp$ be the orthogonal complement of 
$\mathbf{V}_{d,k}$ corresponding to directions of lower variability. 
This distance is also equal to the norm of the difference
between ${S}_J x$ and the average class ``template'' $E ({S}_{J} X_k)$, 
projected in $\mathbf{V}_{d,k}^\perp$:
\begin{equation}
\label{captunsf}
\|{S}_J x - P_{\mathbf{A}_{d,k}}({S}_J x) \| = 
\Big\|P_{\mathbf{V}_{d,k}^\perp} \Big({S}_J x - E ({S}_{J} X_k) \Big) \Big\|~.
\end{equation}
Minimizing the affine space approximation error is thus
equivalent to finding the class centroid $E({S}_J X_k)$ which is the
closest to ${S}_J x$, without taking into account the first $d$ 
principal variability directions. 
The $d$ principal directions of the space $\mathbf{V}_{d,k}$ result
from deformations and from structural variability.
The projection $P_{\mathbf{A}_{d,k}}({S}_J x)$ is
the optimum linear prediction of ${S}_J x$
from these $d$ principal modes.
The selected class has the smallest prediction error.

This affine space selection is effective if
${S}_J X_k - E ({S}_J X_k )$ is well approximated by a projection in a 
low-dimensional space. 
This is the case if realizations of $X_k$ 
are translations and limited deformations of a single template. 
Indeed, the Lipschitz continuity condition implies that
small deformations are linearized by the scattering transform.
Hand-written digit recognition is an example. 
This is also valid
for stationary textures where ${S}_J X_k$ has a small variance,
which can be interpreted as structural variability. 

The dimension $d$ must be
adjusted so that ${S}_J X_k$ has a better
approximation in the affine space
${\bf A}_{d,k}$ than in affine spaces
${\bf A}_{d,k'}$ of other classes $k' \neq k$. This is a model
selection problem, which requires to optimize the dimension $d$ in
order to avoid over-fitting \cite{BirgeMasard}.

The invariance scale $2^J$ must also be optimized. 
When the scale $2^J$ increases, translation invariance increases but it comes
with a partial loss of information which brings the representations
of different signals closer. One can prove \cite{mallat} that for any $x$ and $x'$
\[
\|S_{J+1} x - S_{J+1} x' \| \leq \|S_{J} x - S_{J} x' \| ~.
\]
When $2^J$ goes to infinity, this 
scattering distance converges to a non-zero value.
To classify deformed templates such as hand-written digits, 
the optimal $2^J$
is of the order of the maximum pixel displacements due to translations
and deformations.
In a stochastic framework where $x$ and $x'$ are realizations 
of stationary
processes, 
$S_J x$ and $S_J x'$ converge to the expected
scattering transforms $\overline{S}x$ and $\overline{S}x'$. 
In order to classify stationary processes such as textures, 
the optimal scale is the maximum scale equal to the image width, 
because it minimizes the variance of the windowed scattering estimator.

A cross-validation procedure is used to 
find the dimension $d$ and the scale $2^J$ 
which yield the smallest classification
error. This error is computed on a subset of the training images,
which is not used to estimate the covariance matrix for the PCA calculations.

As in the case of SVM, the performance of the affine 
PCA classifier can be improved by equalizing the descriptor space. 
Table \ref{scat_energy_absortion1} shows that scattering vectors 
have unequal energy distribution along its path variables, in particular
as the order varies. A robust equalization is obtained by re-normalizing 
each $S_J[p] X(u)$ by the maximum 
$\| S_J [p] X_i \| = \Big(\sum_u |S_J [p] X_i(u)|^2 \Big)^{1/2}$
over all training signals $X_i$:
\begin{equation}
\label{l_infy_renorm}
\frac{S_J[p] X(u)}{\sup_{X_i} \| S_J [p] X_i \|}~.
\end{equation}
To simplify notations, we still write
$S_J X$ for this normalized scattering vector.

Affine space scattering models can be interpreted
as generative models computed independently for each class. 
As opposed to
discriminative classifiers such as SVM, they do not estimate 
cross-terms between classes, besides from the
choice of the model dimensionality $d$. Such estimators are
particularly effective for small number of training samples per
class. Indeed, if there are few training samples per class then
variance terms dominate bias errors when estimating 
off-diagonal covariance coefficients between classes
\cite{Bickel}.

An affine space approximation classifier
can also be interpreted as a robust quadratic discriminant classifier obtained
by coarsely quantizing the eigenvalues of 
the inverse covariance matrix. For each class, the eigenvalues of
the inverse covariance are set to 
$0$ in $\V_{d,k}$ and to $1$ in $\V^\perp_{d,k}$, where $d$ is adjusted by
cross-validation. This coarse quantization 
is justified by the poor estimation of covariance eigenvalues from few
training samples. These affine space models will 
typically be applied to distributions of scattering vectors
having non-Gaussian distributions, where a Gaussian Fisher discriminant 
can lead to important errors. 

\subsection{Handwritten Digit Recognition}
\label{handwritten}

The MNIST database of hand-written digits is an example 
of structured pattern classification, where 
most of  the intra-class variability is due to
local translations and deformations. It comprises 
at most $60000$ training samples and $10000$ test samples. 
If the training dataset is not augmented with deformations, 
the state of the art was achieved by deep-learning convolutional networks
\cite{ranzato_cvpr}, deformation models \cite{mnist_deformation},
and dictionary learning \cite{mairal}. These results are improved
by a scattering classifier.

All computations are performed on the reduced cosine scattering
representation described in Section \ref{CosineSca},
which keeps the lower-frequency half of the coefficients.
Table \ref{full_mnist_results} computes classification errors on a fixed set of test
images, depending upon the size of the training set, for different
representations and classifiers. The affine space selection of
section \ref{secaffi} is compared with an SVM 
classifier using RBF kernels,
which are computed using Libsvm \cite{libsvm}, and 
whose variance is adjusted using standard cross-validation
over a subset of the training set.
The SVM classifier is trained with a renormalization which maps all
coefficients to $[-1,1]$. The PCA classifier is trained with
the renormalisation (\ref{l_infy_renorm}).
The first two columns of
Table \ref{full_mnist_results} show that classification 
errors are much smaller with an SVM 
than with the PCA algorithm if applied directly on the image.
The 3rd and 4th columns give the classification error obtained with a PCA
or an SVM classification applied to the modulus of a windowed 
Fourier transform.
The spatial size $2^J$ of the window is optimized with a cross-validation
which yields a minimum error for $2^J = 8$. It corresponds to 
the largest pixel displacements due to
translations or deformations in each class. Removing the complex phase of
the windowed Fourier transform yields a locally invariant representation but
whose high frequencies are unstable to deformations, as explained in 
Section \ref{transdef}. Suppressing this local translation variability
improves the classification rate by a factor $3$ for a PCA and by 
almost $2$ for an SVM.
The comparison between PCA and SVM confirms the fact
that generative classifiers can outperform discriminative classifiers
when training samples are scarce \cite{gen_vs_disc}. 
As the training set size increases, 
the bias-variance trade-off turns in 
favor of the richer SVM classifiers, 
independently of the descriptor.

\begin{figure*}[ht]
\centering
\subfigure[]{
\includegraphics[scale=0.32]{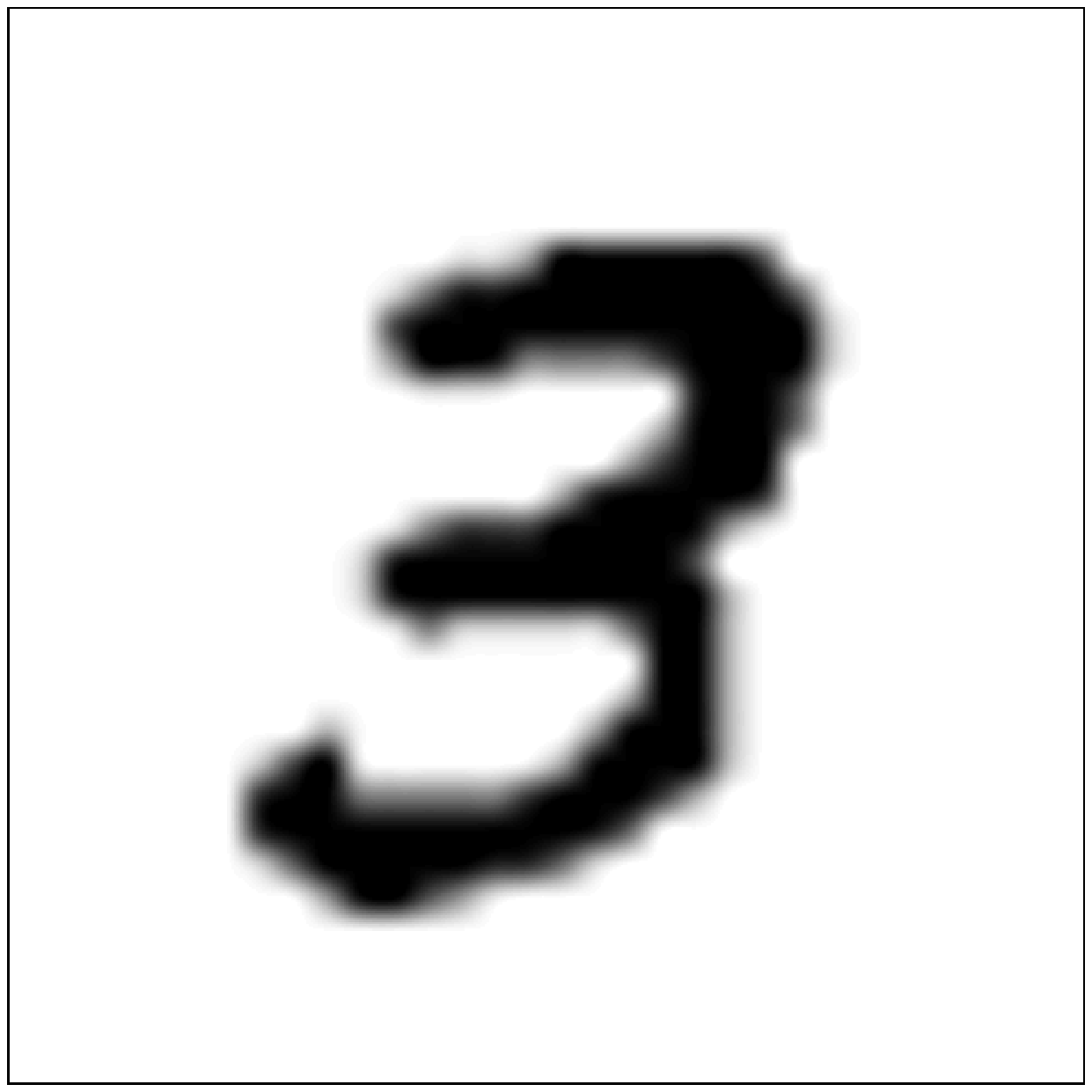}
} 
\subfigure[]{
\includegraphics[scale=0.41,trim=0.5in 0.5in 0.3in 0,clip]{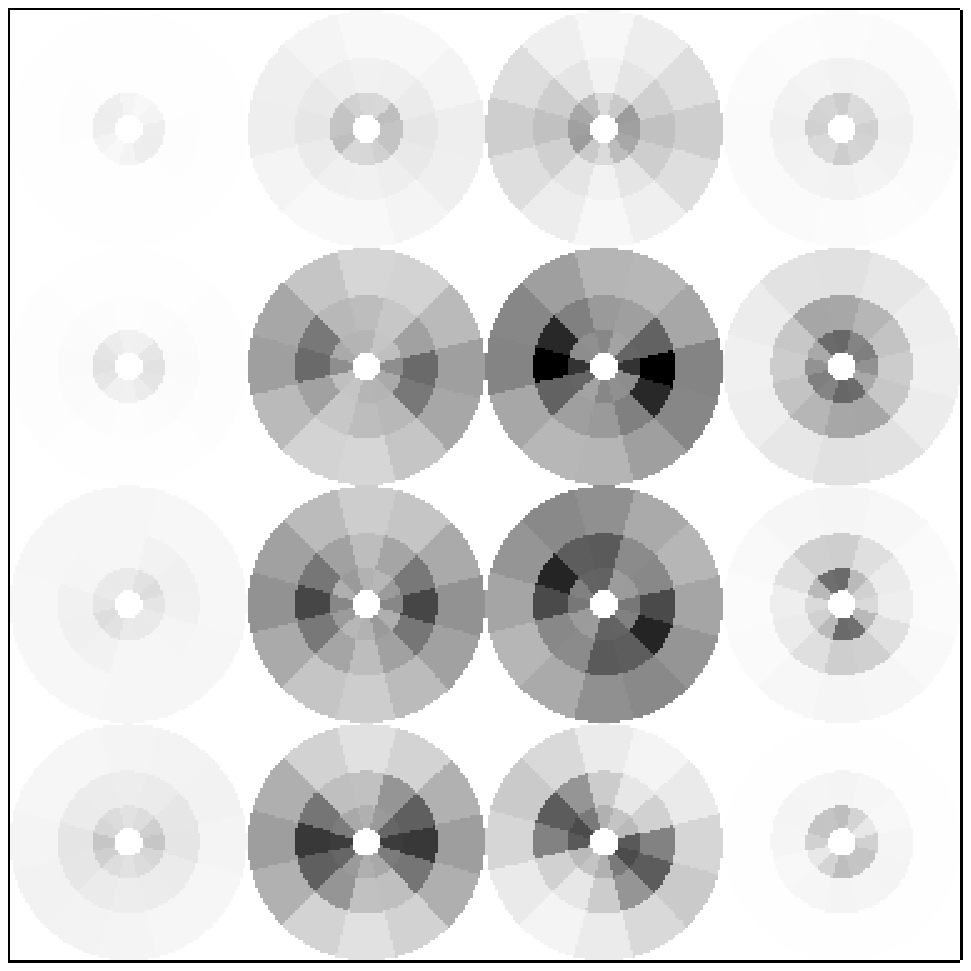}
}
\subfigure[]{
\includegraphics[scale=0.41,trim=0.5in 0.5in 0.3in 0,clip]{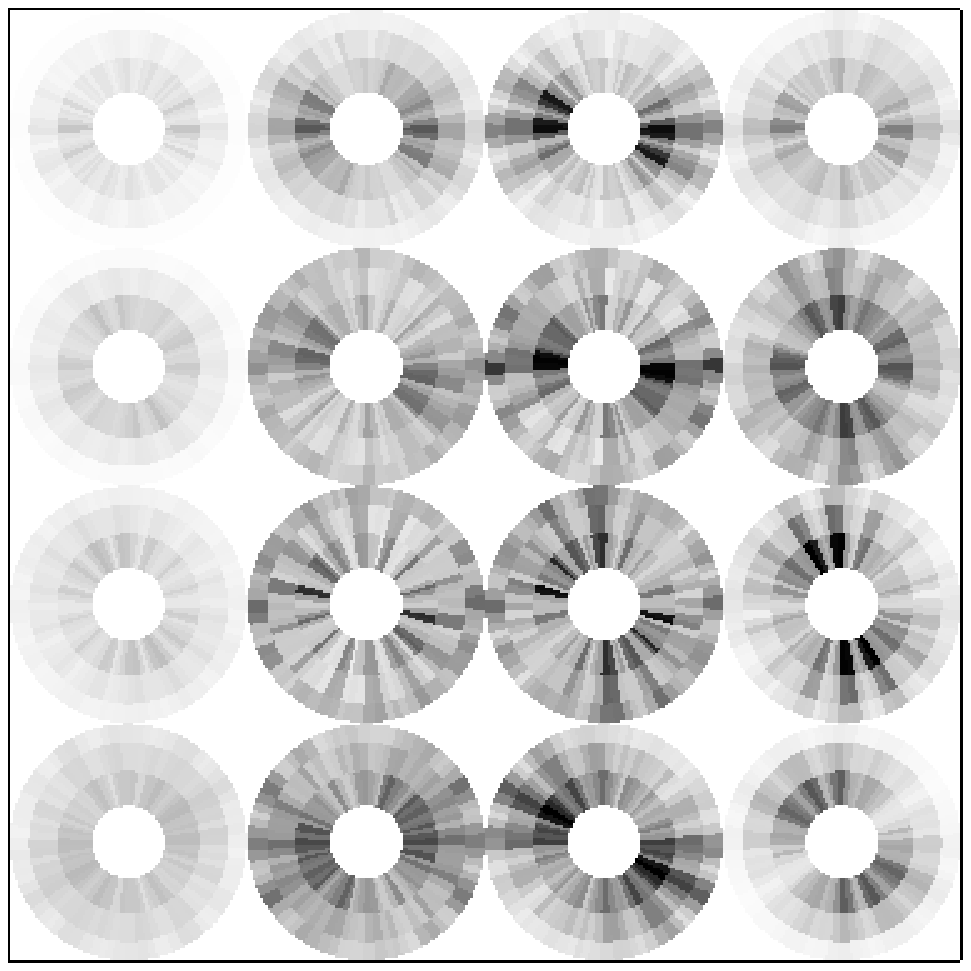}
} \\
\caption{(a): Image $X(u)$ of a digit '3'. 
(b): Array of scattering vectors ${S_J}[p]X(u)$, for $m=1$ and $u$
sampled at intervals $2^J = 8$.
(c): Scattering vectors ${S_J}[p]X(u)$, for $m=2$. }
\label{mnist_scat_analysis}
\end{figure*}

Columns 6 and 8 give the PCA classification result applied to a windowed
scattering representation for $m_{\max} = 1$ and  $m_{\max} = 2$.
The cross validation also chooses $2^J = 8$.
For the digit `3', 
Figure \ref{mnist_scat_analysis} displays the 4-by-4 array of 
normalized scattering vectors.
For each $u = 2^J (n_1,n_2)$ with $1 \leq n_i \leq 4$, the scattering
vector $S_J[p] X(u)$ is displayed for 
paths of length $m=1$ and $m=2$, as circular frequency energy
distributions  following Section \ref{convnetwosec}.

\setcounter{subfigure}{0}


Increasing the scattering order from $m_{\max} = 1$ to $m_{\max} = 2$
reduces errors by about $30$\%,
which shows that second order coefficients
carry important information even at a relatively small scale $2^J = 8$.
However, third order coefficients have a negligible energy and including them
brings marginal classification improvements, 
while increasing computations by an important factor.
As the learning set increases in size, the classification improvement of
a scattering transform increases relatively to
windowed Fourier transform because the classification is able to 
incorporate more high frequency structures, which have deformation
instabilities in the Fourier domain as opposed to the scattering domain. 

%

Table \ref{full_mnist_results} also shows that
below $5\cdot\, 10^3$ training samples, the scattering PCA classifier 
improves results of a deep-learning convolutional networks, which learns
all filter coefficients with a back-propagation algorithm \cite{LeCun}.
As more training samples are available, the
flexibility of the SVM classifier brings an improvement over 
the more rigid affine classifier, yielding a $0.43\%$ error rate
on the original dataset, thus improving upon previous state of the
art methods. 

\begin{table*}[t]
\caption{MNIST classification results.}
\begin{center}
\begin{tabular}{|c |c c | c c| c c| c c | c|}
\hline
Training & \multicolumn{2}{|c|}{$x$}   & \multicolumn{2}{|c|}{Wind. Four.} & \multicolumn{2}{|c|}{Scat. $m_{\max}=1$} &\multicolumn{2}{|c|}{Scat. $m_{\max}=2$} & Conv. \\ 
size  & PCA & SVM & PCA & SVM & PCA & SVM & PCA & SVM  & Net. \\ 
\hline
300& $14.5$        & $15.4$ & $7.35$ & $7.4 $ & $5.7$ & $8$ & $\bf{4.7}$ & $5.6$& $7.18$     \\ 
1000 & $7.2$     & $8.2$   & $3.74$ & $3.74$ & $2.35$ & $4$ & $\bf{2.3}$   & $2.6$& $3.21$   \\
2000 & $5.8$     & $6.5$    &$2.99$ & $2.9$ & $1.7$ & $2.6$ & $\bf{1.3}$ & $1.8$& $2.53$   \\
5000 & $4.9$     & $4$       & $2.34$ & $2.2$ & $1.6$ & $1.6$ & $\bf{1.03}$  & $1.4$ & $1.52$ \\
10000 & $4.55$ & $3.11$ & $2.24$ & $1.65$ & $1.5$ & $1.23$ & $0.88$  & $1$&  $\bf{0.85}$  \\
20000 & $4.25$ & $2.2$   & $1.92$ & $1.15$ & $1.4$ & $0.96$ & $0.79$  &  $\bf{0.58}$& $0.76$  \\
40000 & $4.1$   & $1.7$   & $1.85$ & $0.9$ & $1.36$ & $0.75$ & $0.74$  & $\bf{0.53}$& $0.65$  \\
60000 & $4.3$   &$1.4$    & $1.80$ & $0.8$ & $1.34$ & $0.62$ & $0.7$  & $\bf{0.43}$&  $0.53$  \\
\hline
\end{tabular}
\end{center}
\label{full_mnist_results}
\end{table*}


To evaluate the precision of the affine space model, we compute
the relative affine 
approximation error, averaged over all classes:
\[
\sigma_d^2 = K^{-1} \sum_{k=1}^K \frac
{E ( \|{S}_J X_k - P_{\bf A_{d,k}} ({S}_J X_k) \|^2 )} 
{E (\|{S}_J X_k \|^2)}~.
\]
For any ${S}_J X_k$, we also calculate
the minimum approximation error 
produced by another affine model $A_{d,k'}$ with $k'\neq k$:
$$\lambda_d = \frac{E (\min_{k'\neq k}  \|{S}_J X_k - P_{\bf A_{k',d}} ({S}_J X_k) \|^2)}{E (\|{S}_J X_k - P_{\bf A_{d,k}} ({S}_J X_k) \|^2)}~.$$

\begin{table}[t]
\caption{Values of the dimension $d$ of affine approximation models
on MNIST classification, of the 
intra class normalized approximation error $\sigma^2_d$,
and of the ratio $\lambda_d$ between inter class and intra class 
approximation errors, as a function of the training size.}
\label{MNIST_dimension}
\begin{center}
\begin{tabular}{|c | c c c |}
\hline
Training & $d$ & $\sigma^2_d$ & $\lambda_d$ \\ 
\hline
300 & $5$ & $3 \cdot 10^{-1}$ & $2$\\
5000 &  $100$ & $4 \cdot10^{-2}$ & $3$ \\
40000 & $140$ & $2\cdot 10^{-2}$ & $4 $ \\
\hline
\end{tabular}
\end{center}
\end{table}

For a scattering representation with $m_{\max}=2$, 
Table \ref{MNIST_dimension} gives the dimension $d$ of affine
approximation spaces optimized with a cross validation, 
with the corresponding values of $\sigma_d^2$ and $\lambda_d$. 
When the training set size increases, the model dimension
$d$ increases because there are more samples to estimate each 
intra-class covariance matrix. The approximation model becomes
more precise so
$\sigma_d^2$ decreases and the relative approximation error $\lambda_d$ 
produced by wrong classes increases. This explains the 
reduction of the classification error rate observed in Table \ref{full_mnist_results}
as the training size increases.

The US-Postal Service is another handwritten digit 
dataset, with 7291 training samples
and 2007 test images $16 \times 16$ pixels.
The state of the art is obtained with 
tangent distance kernels \cite{tangent}.
Table \ref{usps} gives 
results obtained with a scattering transform
with the PCA classifier for $m_{\max} = 1,2$. 
The cross-validation sets the scattering scale to $2^J = 8$.
As in the MNIST case,
the error is reduced when going from $m_{\max} = 1$ to $m_{\max} = 2$ but
remains stable for $m_{\max} = 3$. 
Different renormalization strategies can bring marginal improvements
on this dataset. If the renormalization is 
performed by equalizing using the standard deviation
of each component, the classification
error is ${2.3 \%}$ whereas it is $2.6 \%$ if 
the supremum is normalized.



\begin{table}[t]
\caption{Percentage of errors for the whole USPS database.}
\label{usps}
\begin{center}
\begin{tabular}{|c |c| c| c|}
\hline
Tang. & Scat. $m_{\max}=2$ & Scat. $m_{\max}=1$ & Scat. $m_{\max}=2$ \\
Kern. & SVM & PCA & PCA \\ 
\hline
${2.4}$ & $2.7$ & $3.24$ & $2.6~/~{\bf 2.3}$  \\
\hline
\end{tabular}
\end{center}
\end{table}

The scattering transform is stable but not invariant to rotations. 
Stability to rotations is demonstrated over the
MNIST database in the setting defined in \cite{mnist_rotated}. 
A  database with 12000 training samples and 50000 test images
is constructed with random rotations of MNIST digits.
The PCA affine space selection takes into account the rotation
variability by increasing 
the dimension $d$ of the affine approximation space. This is equivalent
to projecting the distance to the class centroid on a smaller 
orthogonal space, by removing more principal components.
The error rate in
Table \ref{MNIST_rotated} is much smaller with a scattering PCA than with
a convolution network \cite{mnist_rotated}. 
Much better results are obtained for a scattering
with $m_{\max} =2$ than with $m_{\max} =1$ because
second order coefficients
maintain enough discriminability despite the removal of a larger
number $d$ of principal directions. In this case,
$m_{\max} = 3$ marginally reduces the error.

\begin{table}[t]
\caption{Percentage of errors on an MNIST rotated dataset \cite{mnist_rotated}.}
\label{MNIST_rotated}
\begin{center}
\begin{tabular}{|c| c| c|}
\hline
Scat. $m_{\max}=1$ & Scat. $m_{\max}=2$ &  Conv. \\
PCA & PCA & Net. \\ 
\hline
$8$ & $\bf{4.4}$ & $8.8$ \\
\hline
\end{tabular}
\end{center}
\end{table}

\begin{table}[t]
\caption{Percentage of errors on scaled and/or rotated MNIST digits}
\label{MNIST_transf}
\begin{center}
\begin{tabular}{|c | c| c|}
\hline
Transformed & Scat. $m_{\max}=1$ & Scat. $m_{\max}=2$  \\ 
Images & PCA & PCA\\
\hline
Without & $1.6$ & $0.8$ \\
Rotation & $6.7$ & $3.3$ \\
Scaling & $2$ & $1$ \\
Rot. $+$ Scal. & $12$ & $5.5$ \\
\hline
\end{tabular}
\end{center}
\end{table}

Scaling invariance is studied by introducing a random scaling 
factor uniformly distributed between $1/\sqrt{2}$ and $\sqrt{2}$.
In this case, the digit `9' is removed from the database as 
to avoid any indetermination
with the digit `6' when rotated. 
The training set has $9000$ samples ($1000$ samples per class).
Table \ref{MNIST_transf} gives
the error rate on the original MNIST database and
 including 
either rotation, scaling, or both in the training and testing samples.
Scaling has a smaller impact on the error rate than rotating digits
because scaled scattering vectors span an invariant linear space of lower
dimension. Second-order scattering 
outperforms first-order scattering, and
the difference becomes more significant when
rotation and scaling are combined, because 
it provides interaction coefficients which are 
discriminative even in presence of 
scaling and rotation variability.

\subsection{Texture Discrimination}
\label{texturesec}

Visual texture discrimination remains an outstanding image processing
problem because textures are realizations of non-Gaussian stationary
processes, which cannot be discriminated using the power spectrum.
Depending on the imaging conditions, textures undergo transformations due
to illumination, rotation, scaling or more complex deformations when
mapped on three-dimensional surfaces. 
The affine PCA space classifier 
removes most of the 
variability of $S_J X - E \{S_J X \}$ within each class.
This variability is due to the residual stochastic variability which decays
as $J$ increases and to variability due to illumination, rotation and
perspective effects.

\begin{figure*}[ht]
\centering
\includegraphics[scale=0.35]{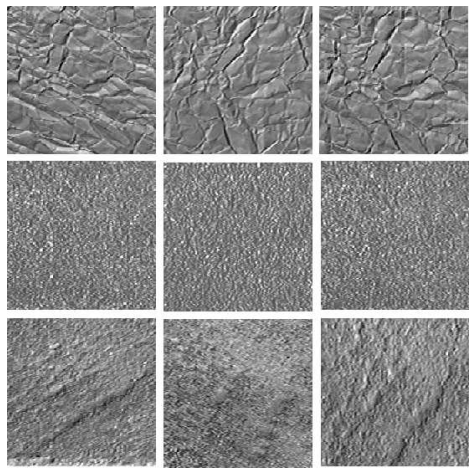}
\caption{Examples of textures from the CUReT database with normalized
mean and variance. Each row corresponds to a different class, showing intra-class variability
in the form of stochastic variability and changes in pose and illumination.
}
\label{curetfig}
\end{figure*}

Texture classification is tested on 
the CUReT texture database \cite{Malik,Zisserman}, which includes
61 classes of image textures of $N = 200^2$ pixels.
Each texture class gives images of
the same material with different
pose and illumination conditions. 
Specularities, shadowing and surface normal 
variations make classification challenging. 
Pose variation requires global rotation and illumination invariance.
Figure \ref{curetfig} 
illustrates the large intra-class variability, after a normalization
of the mean and variance of each textured image.

Table \ref{curet} compares error rates obtained with different classifiers.
The database is randomly split into a training
and a testing set, with 46 training images for each class as in \cite{Zisserman}.
Results are averaged over 10 different splits.
A PCA affine space classifier applied directly on the image
yields a large classification error of $17\%$.
To estimate the Fourier spectrum, 
windowed Fourier transforms are computed over 
half-overlapping windows of size $2^J$,
and their squared modulus is averaged over the whole image.
This averaging is necessary to reduce the spectrum estimator variance, which
does not decrease when the window size $2^J$ increases.
The cross-validation sets the optimal window scale to $2^J = 32$,
whereas images have a width of $200$ pixels. 
The error drops to 1\%. This simple Fourier spectrum
yields a smaller error than previously reported state-of-the-art methods.
SVM's applied to a dictionary of textons yield 
an error rate of 1.53\% \cite{hayman}, whereas
an optimized Markov Random Field model
computed with image patches \cite{Zisserman}
achieves an error of 2.46\%.

\begin{table*}[t]
\caption{Percentage of errors on CUReT for different training sizes.} 
\label{curet}
\begin{center}
\begin{tabular}{|c |c | c| c| c| c| c|}
\hline
Training  & $X$ & Four. Spectr. & Scat. $m_{\max}=1$ & Scat. $m_{\max}=2$ & Textons & MRF  \\
size & PCA & PCA & PCA & PCA & SVM \cite{hayman} & \cite{Zisserman}  \\ 
\hline
46 & $17$ & $1$ & $0.5$ & $\bf{0.2}$ & $1.53$ & $2.4$ \\
\hline
\end{tabular}
\end{center}
\end{table*}

\setcounter{subfigure}{0}
\begin{figure*}[ht]
\centering
\subfigure[]{
\includegraphics[scale=0.25, trim=0 0in 0 0, clip]{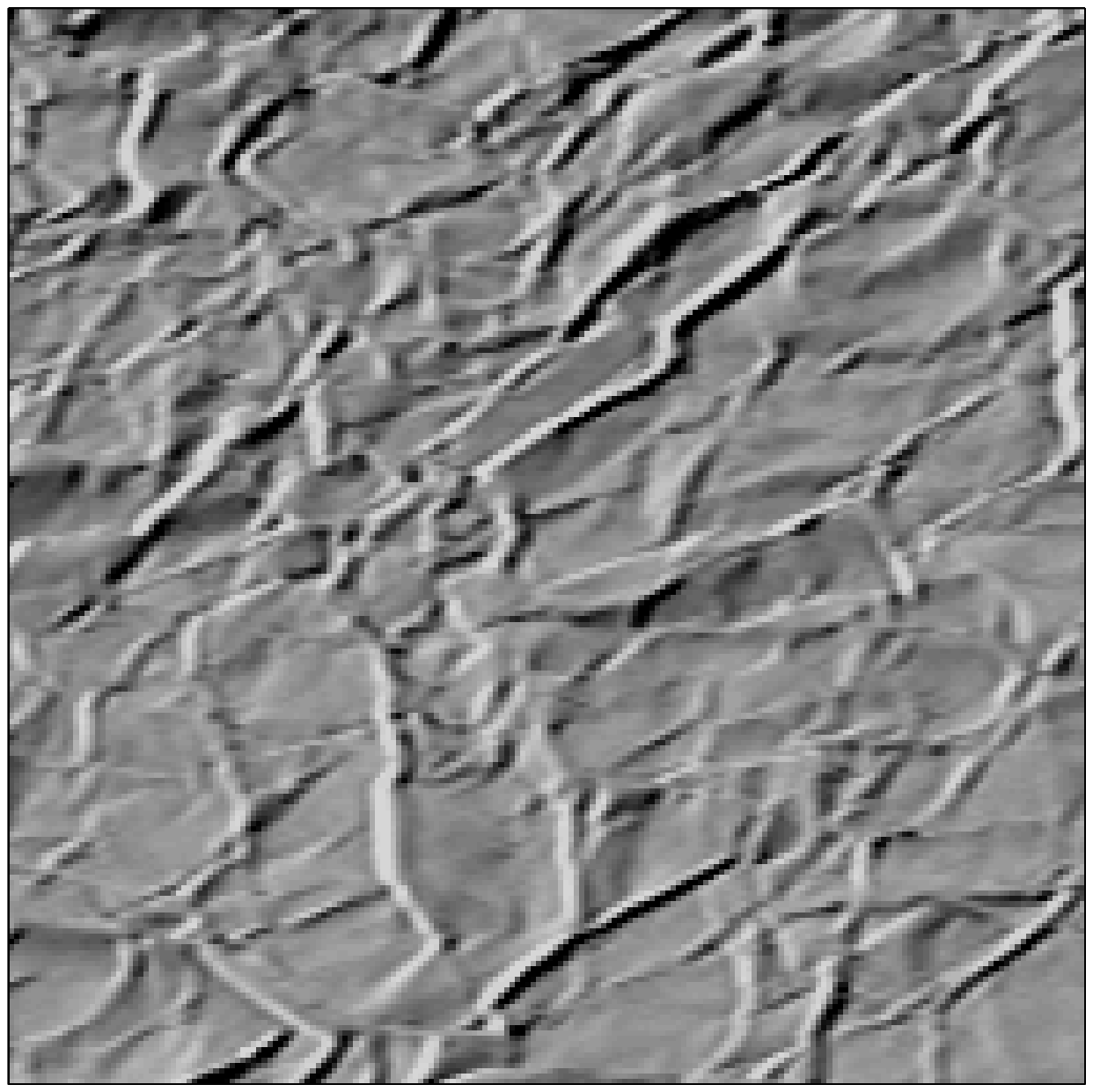}
} 
\subfigure[]{
\includegraphics[scale=0.32,trim=0.75in 0.5in 0.75in 0, clip]{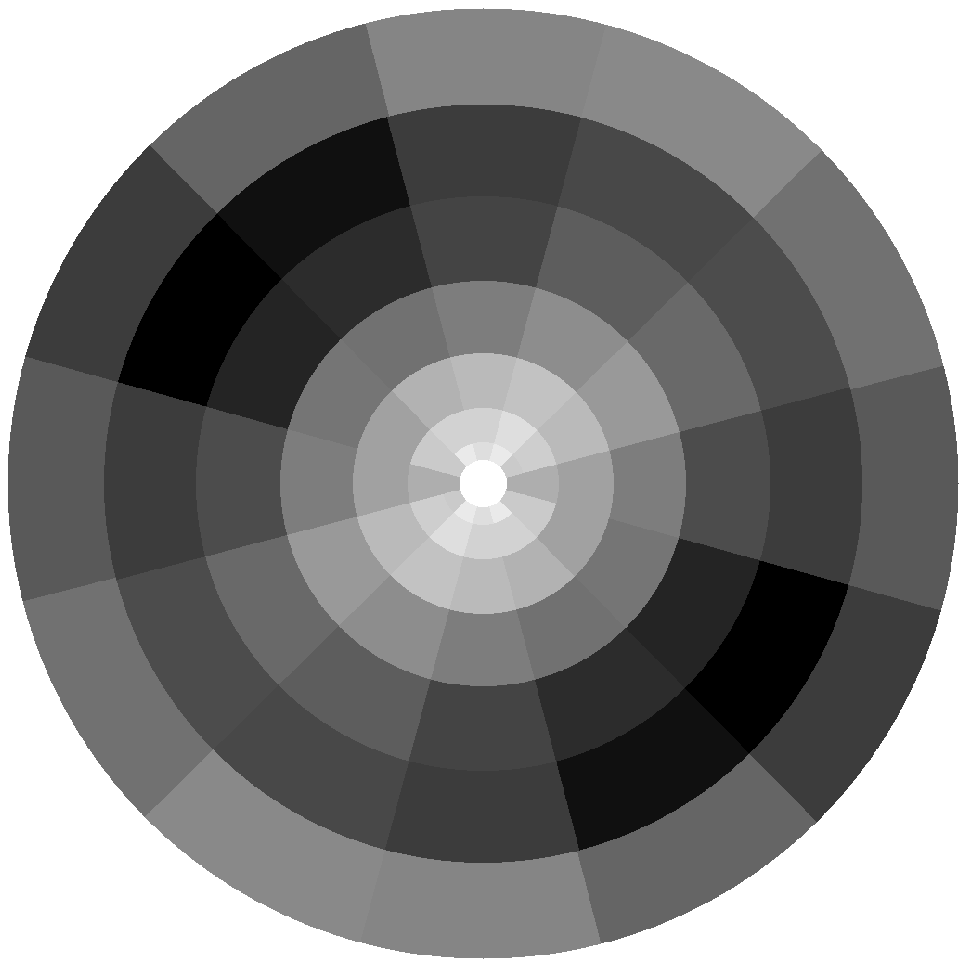}
}
\subfigure[]{
\includegraphics[scale=0.32,trim=0.75in 0.5in 0.75in 0, clip]{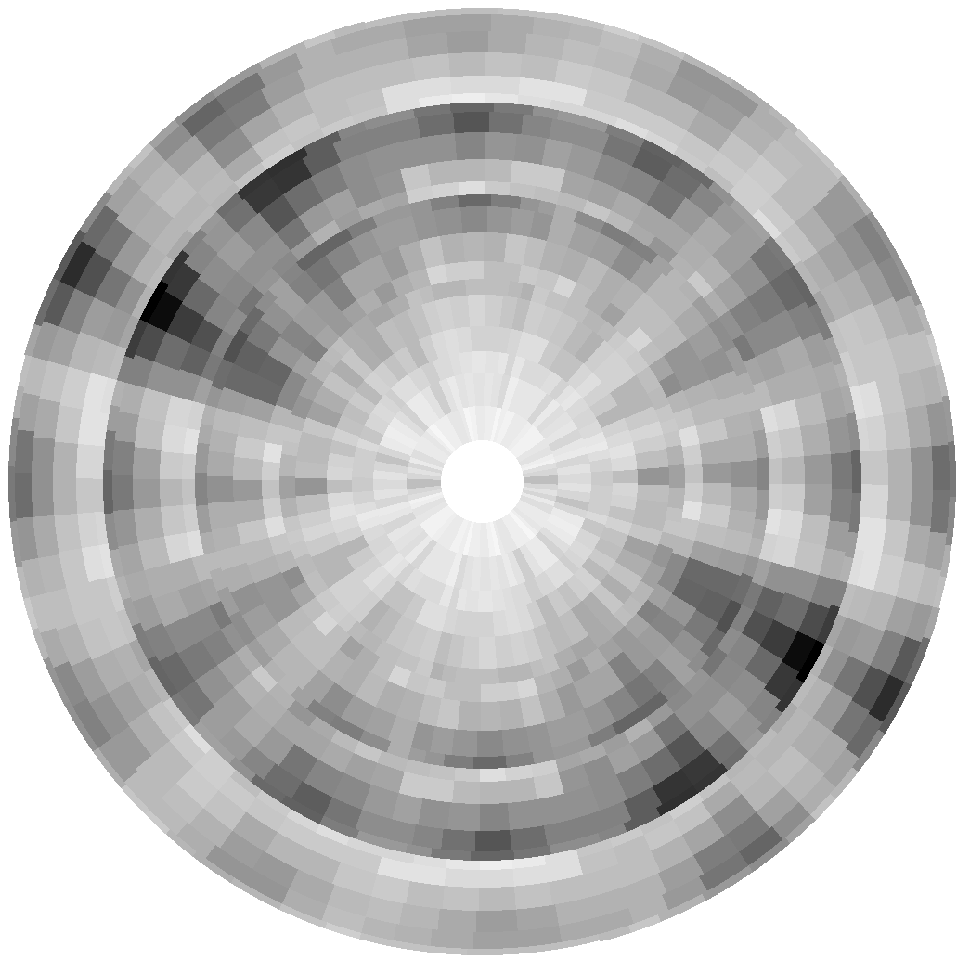}
} 
\caption{ 
(a): Example of CureT texture $X(u)$. (b):
Scattering coefficients $S_J[p]X$, for $m=1$ and $2^J$ equal to the image width.
(c): Scattering coefficients $S_J[p]X(u)$, for $m=2$.}
\label{cure_scat_analysis}
\end{figure*}

For the scattering PCA classifier, 
the cross validation chooses
an optimal scale $2^J$ equal to the image width to reduce the scattering
estimation variance. Indeed, contrarly to a power spectrum estimation,
the variance of the scattering vector decreases when $2^J$ increases.
Figure \ref{cure_scat_analysis} displays the scattering coefficients
$S_J[p] X $ of order $m=1$ and $m=2$ of a CureT textured image $X$.
When $m_{max}=1$, the error drops to 
0.5\%, although first-order scattering coefficients essentially depend
upon second order moments as the Fourier spectrum. This is probably due
to the fact that image textures have a spectrum which typically
decays like $|\omega|^{-\alpha}$. For such spectrum, an estimation over
dyadic frequency bands provide a better bias versus variance trade-off
than a windowed Fourier spectrum \cite{abry}. For
$m_{max}=2$, the error further drops to 0.2\%.
Indeed, scattering coefficients of order $m = 2$ depend upon
moments of order $4$, which are necessary to differentiate
textures having same second order moments as in Figure \ref{norm_textures}.
The dimension of the affine approximation space model is $d=16$,
the intra-class normalized approximation error is
$\sigma^2_d = 2.5 \cdot 10^{-1}$ and 
the separation ratio is $\lambda_d = 3$ for $m_{\max}=2$.

The PCA classifier
provides a partial rotation invariance by removing principal components.
It averages scattering coefficients along path rotation parameters, which
comes with a loss of discriminability. However, 
a more efficient rotation invariant texture classification is obtained
by cascading this translation invariant scattering with a second 
rotation invariant scattering \cite{sifre}. 
It transforms each layer of the translation invariant scattering 
network with new wavelet convolutions along rotation parameters, followed
by modulus and average pooling operators, which are cascaded.
A combined translation and rotation
scattering yields a translation and rotation invariant representation which
is stable to deformations \cite{mallat}.

\section{Conclusion}

A wavelet scattering transform computes a translation invariant representation,
which is stable to deformation, using a deep convolution network 
architecture. The first layer outputs SIFT-type descriptors,
which are not sufficiently informative for large-scale invariance. 
Classification
performance is improved by adding other layers providing 
complementary information.
A reduced cosine scattering transform is at most three times 
larger than a SIFT descriptor and computed with $O(N \log N)$ operations.

State-of-the-art classification results are obtained
for handwritten digit recognition and texture discrimination, 
with an SVM or a PCA classifier.
If the data set has other sources of variability due to the action of other
finite Lie groups such as rotations, then this variability can
be eliminated with an invariant
scattering computed by cascading
wavelet transforms defined on these groups \cite{mallat,sifre}. 

However, signal classes may also
include complex sources of variability that can not be approximated by the
action of a finite group, as in CalTech101 or Pascal databases. 
This variability must be taken into account by unsupervised optimizations
of the representations from the training data.
Deep convolution networks which learn filters from the data 
\cite{LeCun} have the flexibility to adapt to such variability,
but learning translation invariant filters is not necessary. 
A wavelet scattering transform can be used
on the first two network layers, while learning the next layer filters
applied to scattering coefficients.
Similarly, bag-of-features unsupervised 
algorithms \cite{llc,midlevel} applied to SIFT can potentially
be improved upon by replacing SIFT descriptors by
wavelet scattering vectors.


\begin{thebibliography}{99}

\bibitem{abry}
P. Abry, P. Gonçalves, and P. Flandrin, ``Wavelets, spectrum analysis and 1/f processes'', Wavelets and statistics, Lecture Notes in Statistics, 1995.

\bibitem{allassoniere}
S. Allassonniere, Y. Amit, A. Trouve, ``Toward a coherent statistical framework for dense deformable template estimation". Volume 69, part 1 (2007), pages 3-29, of the Journal of the Royal Statistical Society.

\bibitem{bajcsy}
R. Bajcsy and S. Kovacic, ``Multi-resolution elastic matching", Computer Vision Graphics
and Image Processing, vol 46, Issue 1, April 1989.

\bibitem{Bickel}
P. J. Bickel  and E. Levina: ``Covariance regularization by thresholding'', Annals of Statistics, 2008.

\bibitem{BirgeMasard}
L. Birge and P. Massart.  ``From model selection to adaptive estimation." In Festschrift for Lucien Le Cam: Research Papers in Probability and Statistics, 55 - 88, Springer-Verlag, New York, 1997.

\bibitem{Joan}
J. Bruna, ``Operators commuting with diffeomorphisms'', CMAP Tech. Report, 
Ecole Polytechnique, 2012.

\bibitem{midlevel}
Y-L. Boureau, F. Bach, Y. LeCun, and J. Ponce. ``Learning Mid-Level Features For Recognition". In IEEE Conference on Computer Vision and Pattern Recognition, 2010.

\bibitem{Poggio}
J. Bouvrie, L. Rosasco, T. Poggio: ``On Invariance in Hierarchical Models", NIPS 2009.

\bibitem{libsvm}
C. Chang and C. Lin, ``LIBSVM : a library for support vector machines". ACM Transactions on Intelligent Systems and Technology, 2:27:1--27:27, 2011

\bibitem{caltech}
L. Fei-Fei, R. Fergus and P. Perona. ``Learning generative visual models
from few training examples: an incremental Bayesian approach tested on
101 object categories". IEEE. CVPR 2004

\bibitem{curet_rotation}
Z. Guo, L. Zhang, D. Zhang, ``Rotation Invariant texture classification using LBP variance (LBPV) with global matching", Elsevier Journal of Pattern Recognition, Aug. 2009.

\bibitem{tangent}
B.Haasdonk, D.Keysers: ``Tangent Distance kernels for support vector machines", 2002.

\bibitem{hayman}
E. Hayman, B. Caputo, M. Fritz and J.O. Eklundh, ``On the Signiﬁcance of Real-World Conditions
for Material Classiﬁcation", ECCV, 2004.

\bibitem{mnist_ranzato}
K. Jarrett, K. Kavukcuoglu, M. Ranzato and Y. LeCun: ``What is the Best Multi-Stage Architecture for Object Recognition?", Proc. of ICCV 2009.

\bibitem{mnist_deformation} D.Keysers, T.Deselaers, C.Gollan, H.Ney, ``Deformation Models for image recognition", IEEE trans of PAMI, 2007.

\bibitem{mnist_rotated}
H. Larochelle,  Y. Bengio, J. Louradour, P. Lamblin, ``Exploring Strategies for Training Deep Neural Networks", Journal of Machine Learning Research, Jan. 2009.

\bibitem{ponce}
S. Lazebnik, C. Schmid,  J.Ponce.
``Beyond Bags of Features: Spatial Pyramid Matching for Recognizing Natural Scene Categories".
Proceedings of the IEEE Conference on Computer Vision and Pattern Recognition, New York, June 2006, vol. II, pp. 2169-2178.

\bibitem{LeCun}
Y. LeCun, K. Kavukvuoglu and C. Farabet: ``Convolutional Networks and Applications in Vision", Proc. of ISCAS 2010.

\bibitem{Malik}
T. Leung and J. Malik; ``Representing and Recognizing the Visual Appearance of Materials Using Three-Dimensional Textons". International Journal of Computer Vision, 43(1), 29-44; 2001.

\bibitem{slotine}
W. Lohmiller and J.J.E. Slotine ``On Contraction Analysis for Nonlinear Systems'', Automatica, 34(6), 1998.

\bibitem{SIFT}
D.G. Lowe,  ``Distinctive Image Features from Scale-Invariant Keypoints", International Journal of Computer Vision, 60, 2, pp. 91-110, 2004

\bibitem{mallat}
S. Mallat  ``Group Invariant Scattering'', to appear in 
``Communications in Pure and Applied Mathematics'', 2012,
http://arxiv.org/abs/1101.2286.

\bibitem{stephane0}
S. Mallat, ``Recursive Interferometric Representation'',
Proc. of EUSICO conference, Denmark, August 2010.

\bibitem{sifre} S.Mallat , L. Sifre : ``Combined scattering for rotation invariant texture analysis", 
submitted to ESANN, 2012.

\bibitem{mairal}
J. Mairal,  F. Bach, J.Ponce, ``Task-Driven Dictionary Learning", Submitted to IEEE trans. on PAMI, September 2010.

\bibitem{gen_vs_disc} 
A. Y. Ng and M. I. Jordan ``On discriminative vs. generative classifiers: A comparison of logistic regression and naive Bayes'',
in Advances in Neural Information Processing Systems (NIPS) 14, 2002.

\bibitem{perrinet}
L. Perrinet, ``Role of Homeostasis in Learning Sparse Representations", 
Neural Computation Journal, 2010.

\bibitem{simoncelli}
J.Portilla, E.Simoncelli, ``A Parametric Texture model based on joint statistics of complex wavelet coefficients", IJCV, 2000.

\bibitem{ranzato_cvpr}
M. Ranzato, F.Huang, Y.Boreau, Y. LeCun: ``Unsupervised Learning of Invariant Feature Hierarchies with Applications to Object Recognition", CVPR 2007.

\bibitem{Zeevi}
C. Sagiv, N. A. Sochen and Y. Y. Zeevi, "Gabor Feature Space Diffusion via the Minimal Weighted Area Method'', Springer Lecture Notes in Computer Science, Vol. 2134, pp. 621-635, 2001.

\bibitem{Scholkopf} 
B. Scholkopf and A. J. Smola, ``Learning with Kernels'', MIT Press, 2002.

\bibitem{Soatto} S.Soatto, ``Actionable Information in Vision", ICCV, 2009.

\bibitem{DAISY}
E. Tola, V.Lepetit, P. Fua, ``DAISY: An Efficient Dense Descriptor Applied to Wide-Baseline Stereo", IEEE trans on PAMI, May 2010.

\bibitem{trouve} A. Trouve, L. Younes, ``Local Geometry of Deformable Templates", SIAM Journal on Mathematical Analysis. 2005. Volume: 37, Issue: 1.

\bibitem{Zisserman}
M.Varma, A. Zisserman: ``A Statistical Approach To Material Classification Using Image Patch Exemplars". IEEE Trans. on PAMI, 31(11):2032--2047, November 2009. 

\bibitem{waldspurger} I. Waldspurger, S. Mallat ``Recovering the phase of a complex wavelet transform", CMAP Tech. Report, Ecole Polytechnique, 2012. 

\bibitem{llc}
J.Wang, J.Yang, K.Yu, F.Lv, T.Huang, Y.Gong, ``Locality-constrained Linear Coding for Image Classification", CVPR 2010.

\end{thebibliography}
\end{document}